\documentclass[11pt]{article}

\usepackage[final]{acl}

\usepackage{times}
\usepackage{latexsym}

\usepackage[T1]{fontenc}

\usepackage[utf8]{inputenc}

\usepackage{microtype}

\usepackage{inconsolata}

\usepackage{graphicx}

\usepackage{quoting}
\usepackage{booktabs}
\usepackage{multirow}
\usepackage{makecell}

\usepackage{nicematrix}
\usepackage{tabularray}
\usepackage{pifont}
\usepackage{tikz}
\newcommand{\tr}[1]{\textcolor{red}{#1}}
\usepackage{CJKutf8}
\usepackage{amsmath}
\usepackage{amssymb}   

\usepackage{tcolorbox}
\usepackage{listings}
\usepackage{xcolor}
\definecolor{promptbg}{gray}{0.95}
\usepackage{fancyvrb}
\definecolor{lightblue}{RGB}{46,110,187}
\definecolor{darkred}{RGB}{150,38,31}
\definecolor{darkgreen}{rgb}{0.0, 0.5, 0.0}
\definecolor{blue}{HTML}{3572EF}
\newcommand{\tg}[1]{\textcolor{darkgreen}{#1}}
\newcommand{\tb}[1]{\textcolor{blue}{#1}}
\usepackage{lipsum}

\usepackage{arydshln}
\usepackage{subcaption}  
\usepackage{enumitem}

\title{Exploring Concreteness Through a Figurative Lens}

\author{Saptarshi Ghosh \and Tianyu Jiang \\
  University of Cincinnati \\
  \texttt{ghosh2si@mail.uc.edu, tianyu.jiang@uc.edu} \\}

\begin{document}
\maketitle
\begin{abstract}

Static concreteness ratings are widely used in NLP, yet a word’s concreteness can shift with context, especially in figurative language such as metaphor, where common concrete nouns can take abstract interpretations. While such shifts are evident from context, it remains unclear how LLMs understand concreteness internally. We conduct a layer-wise and geometric analysis of LLM hidden representations across four model families, examining how models distinguish literal vs. figurative uses of the same noun and how concreteness is organized in representation space. We find that LLMs separate literal and figurative usage in early layers, and that mid-to-late layers compress concreteness into a one-dimensional direction that is consistent across models. Finally, we show this geometric structure is practically useful: a single concreteness direction supports efficient figurative-language classification and enables training-free steering of generation toward more literal or more figurative rewrites.
\end{abstract}

\section{Introduction}

Concreteness has been widely studied in psychology, linguistics, and natural language processing (NLP). The degree of concreteness reflects the extent to which a word denotes a perceptible physical entity. Highly concrete words denote objects that can be directly experienced through the senses (e.g., \textit{apple}, \textit{chair}, \textit{wood}). In contrast, words referring to intangible or abstract notions are considered less concrete (e.g., \textit{justice}, \textit{idea}, \textit{freedom}). In linguistic studies, the concreteness of a lexical item is closely tied to whether its interpretation is literal or figurative~\citep{lakoff1980}. While figurativity and concreteness are not equivalent, figurative expressions often involve applying a typically concrete word in a non-physical or non-literal sense. Consider the words denoted in bold in the following sentences:
\begin{quoting}[font={itshape, raggedright, noindent}, leftmargin=1em, rightmargin=0in]
(a) There is a small \textbf{window} of opportunity.

(b) The diary opened a \textbf{door} into his past.

(c) The \textbf{church} joined the movement.
\end{quoting}

Although \textit{window}, \textit{door}, and \textit{church} each denote tangible entities in their canonical literal uses, the above sentences illustrate idiomatic (a), metaphorical (b), and metonymic (c) usage respectively. The intended objects are not directly tied to their physical meanings and refer to something else in the \textit{figurative context}. Recent work highlighted this connection between concreteness and figurativity, showing that figurative usage is often associated with shifts in a word's concreteness~\citep{chakrabarty-etal-2021-mermaid, stowe-etal-2021-metaphor}. Thus, the same lexical item can shift along the concreteness spectrum depending on its contextual interpretation.

In existing NLP work, concreteness is typically treated as a lexical semantic property, most notably through large-scale human annotations such as \citet{concreteness}, which provides static concreteness ratings for words in isolation and have become a foundational reference across the field. Subsequent studies leverage these norms to predict concreteness from word-level embeddings~\citep{charbonnier-wartena-2019-predicting, tater-etal-2022-concreteness, wartena-2024-estimating}, consistently showing that contextualized embeddings reflect concreteness shifts between literal and figurative usage.

However, these studies primarily focus on prediction performance or embedding quality, and do not explicitly examine the internal computation and explainability of contextual concreteness in LLMs. Hence, we still lack a clear understanding of where and how concreteness is processed inside modern LLMs, particularly in relation to figurative text. Mainly, the following questions remain underexplored: (i)~\textit{Which layers of an LLM encode and contextualize concreteness?} (ii)~\textit{Is the concreteness distinction represented along a dedicated geometric subspace in the model’s embedding space?} and (iii) \textit{Can this geometric subspace of concreteness be used for steering?}

Addressing these questions, our work presents a systematic investigation by analyzing the internal representation of concreteness in terms of figurative text across layers and model architectures. First, we use a prompt-based method to generate a token carrying information about the contextual concreteness of a term. Using this, we find layer-wise correlation between the token embedding and concreteness scores from \citet{concreteness}. Our results show that concreteness is encoded in the early layers. We also use a synthetic dataset where a term is used in a pair of high and low concrete sense and show that the early layers distinguish the high and low-concrete usage of the word. Second, we analyze the \textit{internal geometric representation} of concreteness across layers. We compute DiffMean vectors~\citep{diffmean} of high and low concrete terms and create a global concreteness axis using SVD, which we then use to classify sentences. We identify that \textit{later layers} compress concreteness into a one-directional subspace. We exploit this geometry to classify and generate figurative text by showing that the one-directional subspace representation derived from the concreteness axis rivals a full dimensional supervised classifier in figurative language detection. Third, we use this geometry to steer hidden states and show that the concreteness axis can causally shift the figurative expression of the generated output without parameter updates. We conduct large-scale experiments using 25,000 sentences extracted from Wikipedia and evaluate four major model families---Llama-3.1-8B~\citep{grattafiori2024llama3herdmodels}, Qwen3-8B~\citep{yang2025qwen3technicalreport}, Gemma2-9B~\citep{gemma_2024}, and GPT-OSS-20B~\citep{openai2025gptoss120bgptoss20bmodel}, ensuring the generalization of our findings. Our code is publicly available.\footnote{\url{https://github.com/cincynlp/concreteness-interpretability}} In summary, our findings are:
\begin{enumerate}[itemsep=2pt, topsep=4pt]
    \item \textbf{Layer-wise representation of concreteness.} We find that \textit{early layers} of an LLM encode concreteness of a term and distinguish between its high and low concrete use.
    \item \textbf{Geometric subspace of concreteness.} We show that concreteness is compressed into a largely \textit{one-directional subspace in later layers}, revealing a shared geometric organization across LLMs.
    \item \textbf{Geometry-guided figurative language control.} We demonstrate that a one-directional concreteness axis supports both lightweight figurative classification and causal steering of generation, enabling representation-driven control over figurative expression.
\end{enumerate}

\section{Related Work}

Concreteness has long been a focus in psycholinguistics~\citep{lewis}. Early studies showed that the concreteness of a word's meaning in context emerges and changes depending on associative richness and imageability~\citep{Schwanenflugel_Shoben_1983, Barsalou_1999}. Psycholinguistic research shows concrete words are processed faster and more accurately than abstract words in a variety of tasks, a phenomenon known as the \textit{concreteness effect}~\citep{JESSEN2000103, WestHolcomb2000ConcreteAbstractERP, Montefinese2025ConcretenessERP}. Human-generated resources confirm that concreteness forms a continuous semantic dimension rather than a binary category~\citep{concreteness}. In English, the most widely used resource is the \citet{concreteness} dataset of 40,000 words, where each word is rated on a scale of 1-5 rated by over 4,000 human annotators, making this dataset the standard for word-level concreteness. Similar concreteness norms have been extended to other languages, albeit in more limited scope, such as French~\citep{Bonin2018FrenchConcreteness}, Spanish~\citep{Guasch2016SpanishNorms}, and Italian~\cite{Montefinese2023CONcreTEXT, puccetti-etal-2024-abricot}. \citet{Muraki2022MWEConcreteness} collected concreteness judgments for 62k English multiword expressions. 

While static concreteness norms are invaluable, they have inherent limitations~\citep{troche2017}, because many words can be concrete or abstract depending on usage~\citep{frassinelli-etal-2017-contextual}. Generally concreteness of words in isolation highly correlates with that of words in context~\citep{Montefinese2023CONcreTEXT}. But in figurative language, the concreteness of a word significantly alters from its static concreteness~\citep{lakoff1980, Holyoak2018MetaphorReview, Lai2019ActionMetaphorERP, MON2021104285}. Several works have previously leveraged this property of concreteness to identify figurative language such as metaphors and idioms~\citep{tsvetkov-etal-2013-cross, tsvetkov-etal-2014-metaphor, beigman-klebanov-etal-2015-supervised, hall-maudslay-etal-2020-metaphor}. Concreteness has also been extended to multimodal settings, where studies show that it plays a crucial role in shaping how concepts are in figurative depictions such as metaphor~\citep{hessel-etal-2018-quantifying, SU2021166} and metonymy~\citep{ghosh-etal-2026-computational}.

With the advent of powerful LLMs, researchers have studied language models behavior with respect to concreteness. Prior work has shown that contextual concreteness can be predicted using machine learning techniques~\citep{tater-etal-2022-concreteness}. \citet{wartena-2024-estimating} showed that dynamic concreteness estimated from BERT-based contextual embeddings had a very high correlation with human judgment, although the study was limited to encoder only models~\citep{devlin-etal-2019-bert, zhuang-etal-2021-robustly}. The recent work of \citet{Kewenig2025AMT} exemplifies this trend, showing that textual context, visual features and emotional cues to automatically generate concreteness ratings. Interpretability studies have shown that concreteness can be identified as a direction in the word embedding space~\citep{wartena-2022-geometry}. Recent studies demonstrate that the human-to-model alignment is substantially driven by concreteness~\citep{iaia2025representationalalignmenthumanslanguage}. In spite of these advancements, there exists a lack of explainability as to how current LLMs understand concreteness. We do not exactly know which layers, or the subspace geometry that is responsible for distinguishing if a term is used in a high or low concrete sense in the given context. Our study fills this gap by providing insight into model behavior across different LLMs.

\section{Layer-wise Concreteness Representation}
\label{Prompt-based Contextual Probing}

In this section, we discuss the datasets and methodology. We then show which layers of the LLM are responsible for encoding concreteness and differentiating high concrete (literal) usage of a word from its low concrete (figurative) sense.

\subsection{Datasets}

\noindent \textbf{Wikipedia Extraction.} In this work, we restrict our analysis to nouns for three reasons: (i) nouns are the primary carriers of referential meaning in figurative text, and concreteness is traditionally defined over nouns; (ii) nouns frequently exhibit systematic sense shifts between concrete and abstract usage (\textit{root, window, door}), making them ideal for studying figurative contextual variation; (iii) nouns in \citet{concreteness} span a diverse range of concreteness values, enabling controlled evaluation across literal and figurative contexts. 
The statistics from the \citet{concreteness} concreteness dataset further support our decision to focus on nouns. As shown in Table~\ref{table:brysbaert_stat}, nouns exhibit higher variability in concreteness compared to adjectives and verbs. This suggests that nouns span a broader range from concrete to abstract meanings, while verbs and adjectives are more concentrated.

\begin{table}[t]
    \centering
    \resizebox{\linewidth}{!}{
    \setlength{\tabcolsep}{12pt}
    \begin{tabular}{lccc}
    \toprule
    \textbf{POS} & \textbf{Mean} & \textbf{SD} & \textbf{Count}\\
    \midrule
     Noun & 3.52 & 1.01 & 19,056 \\
     Verb & 2.92 & 0.76 & 5,369 \\
     Adjective & 2.49 & 0.51 & 6,112 \\
     \bottomrule
    \end{tabular}
    }
\caption{Statistics of the different parts-of-speech identified from \citet{concreteness}, showing mean concreteness score (Mean), standard deviation (SD), and number of samples (Count).}
\label{table:brysbaert_stat}
\end{table}

In total, we identify 19,056 nouns in the database using spaCy. We then extract \textit{25,000 sentences from Wikipedia dumps} containing these target nouns.\footnote{{https://huggingface.co/datasets/wikimedia/wikipedia}} Extraction proceeds in two stages. First, we collect an initial pool of 50,000 sentences, enforcing a cap of 20 occurrences per noun to maintain lexical diversity. Then, we downsample this pool to 25,000 sentences while preserving a broad distribution of concreteness values across nouns. This ensures that the resulting dataset is both lexically diverse and semantically balanced, providing robust supervision for contextual concreteness probing.

\begingroup
\renewcommand\baselinestretch{0.99}
\begin{table*}[t]
    \NiceMatrixOptions {
    custom-line = {
       command = dashedmidrule ,
       tikz = { dashed } ,
       total-width = \pgflinewidth + \aboverulesep + \belowrulesep ,
     } }
    \centering
    \small
    \resizebox{\textwidth}{!}{
    \begin{NiceTabular}{|p{0.5\linewidth}|p{0.5\linewidth}|}
    \toprule
    \multicolumn{1}{c}{\bf High Concreteness} & \multicolumn{1}{c}{\bf Low Concreteness} \\
    \midrule
    The \tr{\textbf{roof}} was damaged during the storm and needed repairs & The new policy put a \tb{\textbf{roof}} on employee salaries to prevent overpayment. \\ 
    \midrule
    The rusty \tr{\textbf{chain}} bound the old gate shut. & The \tb{\textbf{chain}} of events led to the company's downfall. \\ 
    \midrule
    He climbed the \tr{\textbf{ladder}} to reach the top shelf. & She's been climbing the corporate \tb{\textbf{ladder}} for years. \\ 
    \midrule
    The \tr{\textbf{window}} on the second floor was broken. & The company's new product launch provided a brief \tb{\textbf{window}} into their future plans. \\ 
    \bottomrule
    
    \end{NiceTabular}
    }
    \caption{Example of synthetic GPT-5.1 generated sentences.}
    \label{tab:dataset_example}
\end{table*}
\endgroup

\noindent{\textbf{Synthetic Dataset Creation.}} To enable fine-grained and controlled evaluation, we construct a synthetic contrastive dataset using GPT-5.1. We first sample 600 highly concrete nouns (concreteness score > 4.5) from \citet{concreteness}. For each noun, GPT-5.1 is prompted to generate two different sentences: (i) one where the noun is used in its normal high concrete (literal) sense, and (ii) one where it is used in a low concrete (figurative) sense. This results in 600 sentence pairs, 1,200 sentences total. To verify the generated sentences, two human annotators independently judged whether each sentence used the target noun at the expected level of concreteness. The annotators validated 600 literal sentences and 572 figurative sentences as correct. This portrayed GPT-5.1's strong ability in understanding concreteness and generation. The inter-annotator raw agreement is 93.7\%. Full details about the prompts and annotation are given in Appendix~\ref{synthetic_data_generation}. We randomly sampled 500 pairs from 572, making the synthetic dataset. Table~\ref{tab:dataset_example} provides representative examples from this synthetic corpus. The motivation for using this synthetic dataset is that we want to analyze how LLMs uncover \textit{contextual concreteness} in figurative usage. 

\subsection{Concreteness Token}

The findings of \citet{Montefinese2023CONcreTEXT} and \citet{wartena-2024-estimating} showed that concreteness of words in isolation highly correlates with that of words in context. However, it is well established that in figurative text, the concreteness of a term highly deviate from its static norm~\citep{lakoff1980}. Previous concreteness probing methods typically extract embeddings of the target word in context~\citep{wartena-2024-estimating}. However, in decoder-only LLMs, such embeddings are limited by left-to-right processing and may not fully reflect contextual or figurative interpretations (e.g., ``\textit{The \textbf{chain} of events led to his downfall}''). As a result, we use a prompt-based approach.

\noindent \textbf{Methodology.} To recover contextual semantics, we provide the model with the sentence containing the target noun, asking for its concreteness: 

\begin{quote}\small\ttfamily
Sentence: [sentence]\\
On a scale of 1 to 5 (5 being the highest), in the context of the sentence,
what is the concreteness of the word ``[target\_word]''?
\end{quote}
\vspace{-0.5em}

\noindent We adopt two complementary approaches using this prompt: (i) allowing the model to generate an explicit numerical estimate (Gen), and (ii) extracting the hidden states of the last token representation and train a regression model to predict the concreteness score (Tok). For comparison, we also predict concreteness scores without the context by removing the sentence and providing just the target word to the model. The prompt reads ``\textit{On a scale of 1 to 5 (5 being the highest), what is the concreteness of the word [target\_word].}'' In both cases, we compute and report the Pearson correlation between the predicted score and human-rated concreteness values from \citet{concreteness}.  

\noindent The hidden representations of the last token encode the contextual interpretation of the noun, enabling concreteness probing in generative architectures where direct noun embeddings are insufficient. We extract the hidden states of the last token across all layers. We train a multi-layer perceptron (MLP) regression model using an 80-20 train-test split and predict the concreteness score of the target noun (More details regarding MLP settings in Appendix~\ref{token_generation_prompt}). This setup enables us to study how concreteness is encoded across layers, and how it differs with and without context. Since human concreteness ratings are defined over individual words without context, we expect the without context setting to have a higher correlation.

\begin{table}[t]
\centering
\resizebox{0.98\linewidth}{!}{
\begin{tabular}{lcccc}
\toprule
 & \multicolumn{2}{c}{\textbf{With Context}} & \multicolumn{2}{c}{\textbf{W/o Context}} \\
\cmidrule(lr){2-3} \cmidrule(lr){4-5}
\textbf{Model} & \textbf{Gen} & \textbf{Tok} &  \textbf{Gen} & \textbf{Tok} \\
\midrule
Llama-3.1-8B & 0.66 & \textbf{0.88} & 0.70 & \textbf{0.98}  \\
Qwen3-8B & 0.60 & \textbf{0.87} & 0.65 & \textbf{0.98}  \\
Gemma2-9B & 0.64 & \textbf{0.92} & 0.68 & \textbf{0.98} \\
GPT-OSS-20B & 0.58 & \textbf{0.82} & 0.63 & \textbf{0.98} \\
\bottomrule
\end{tabular}
}
\caption{Pearson correlation between the human concreteness ratings and model predictions using two methods: Gen---numeric rating generated by the model; Tok---score derived from last token hidden states.}
\label{table:prompt_correlation_baseline}
\end{table}

\begin{figure}[t]
    \centering
    \includegraphics[width=0.95\linewidth]{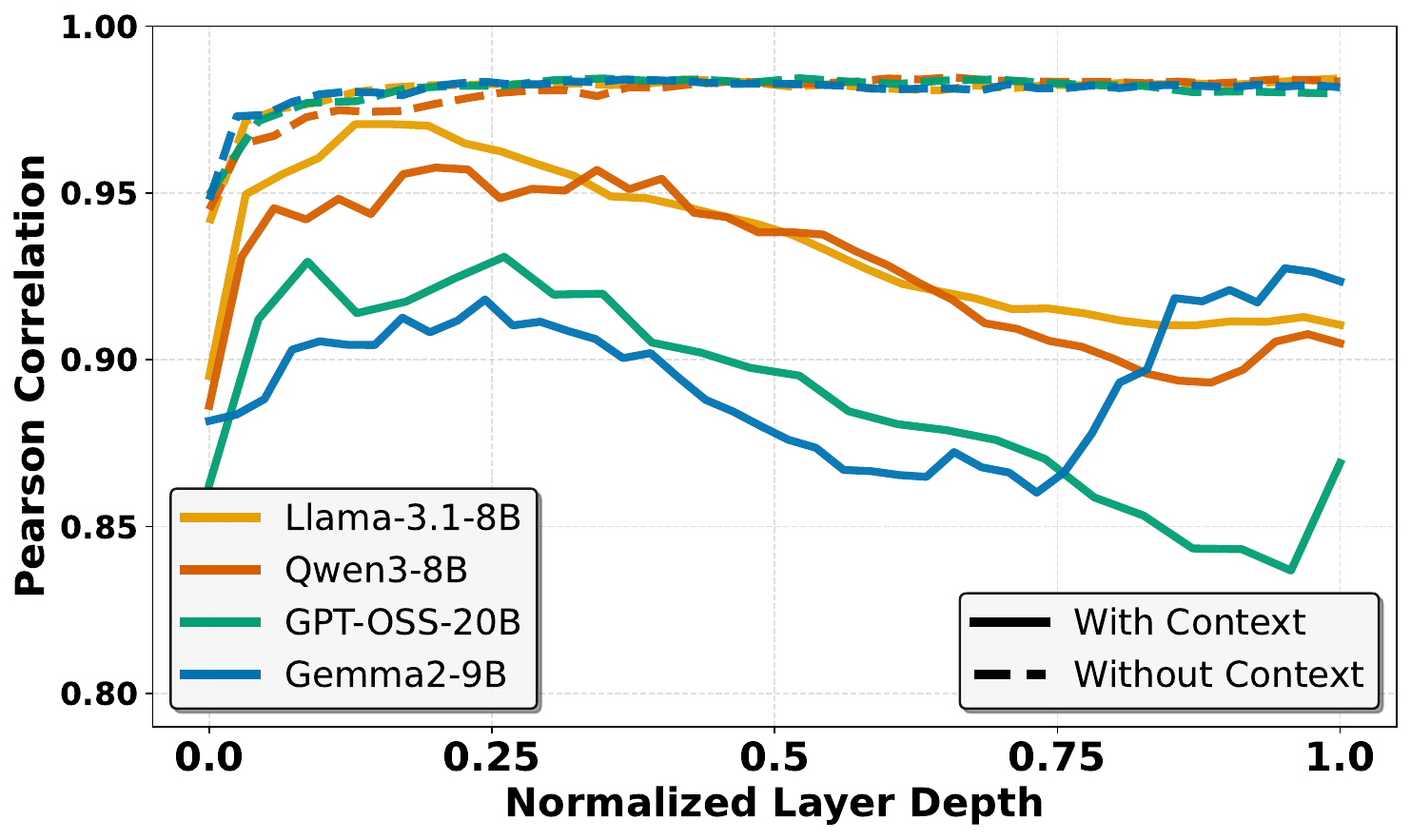}
\caption{Layer-wise Pearson correlation between static human concreteness ratings and model predictions using the last token representation, with and without sentence context, across four LLMs. Contextual input yields an early peak followed by a later decline, indicating increasing deviation from static norms as deeper layers incorporate contextual interpretation.}
\label{fig:exp_1_correlation}
\end{figure}

\noindent \textbf{Results.} Table~\ref{table:prompt_correlation_baseline} reports the Pearson correlation between human-rated concreteness scores and model predictions obtained from the two methods, with and without the sentence. When directly prompted to output a concreteness rating, the models achieve reasonable correlation with human judgments, while predictions derived from the hidden states of the last token representation show consistently higher correlation, demonstrating that the last-token representation encodes substantial information about the concreteness of the target noun.

Figure~\ref{fig:exp_1_correlation} presents the layer-wise (normalized across models) correlation between token representations and human concreteness ratings across all models. As we expected, the correlation score for the without context setting is higher than contextual setting. In both settings, correlation is already high in the earliest layers, suggesting that \textit{concreteness is encoded in the very initial layers}. In without context settings, where the target noun is presented in isolation, this high correlation persists across layers. In contrast, when sentence context is provided, the correlation exhibits a distinct pattern: it peaks in the early layers and gradually decreases in later layers.  This pattern suggests that early layers capture lexical concreteness, while later layers increasingly incorporate sentence-level context such as figurative interpretations.

\subsection{Layers Discriminating Literal \& Figurative Usage}

A shift in a term's concreteness is often brought by the figurative use of that term~\citep{stowe-etal-2021-metaphor}. If concreteness is encoded in the early layers, does that also mean these layers are responsible for distinguishing literal (high-concrete) and figurative (low-concrete) usage of a term? Here, we answer this question with a controlled experiment on the synthetic dataset which contains high and low concrete usage of the same noun. 

Using the 25,000 Wikipedia-sentence corpus, we first extract the hidden states of the last token representation using our prompt, and train an MLP (same settings as the previous experiment) to predict concreteness for each layer-$l$. We then evaluate the trained regression model on the \textit{synthetic dataset}, which contains pairs where the same noun appears in either a literal (highly concrete) usage (e.g., ``the \textit{window} was broken'') or figurative (less concrete) usage (e.g., ``the \textit{window} of opportunity has passed'').

For each layer~$l$, we compute the difference between the predicted contextual concreteness score $C_{\text{pred}}^{(l)}$ and the static human score $C_{\text{static}}$ from \citet{concreteness}, and report the mean difference:
\[
\delta_{\text{mean}}^{(l)} \;=\; \frac{1}{n}\sum_{i=1}^{n} \left( C_{\text{pred},i}^{(l)} - C_{\text{static},i} \right),
\]
where $n$ denotes the number of sentences (500 for each literal and figurative). 
The \emph{magnitude} of $\delta_{\text{mean}}^{(l)}$ (i.e., $|\delta_{\text{mean}}^{(l)}|$) reflects how far the predicted contextual concreteness deviates from the static score. The \textit{sign} indicates the direction of the shift (higher vs.\ lower than the static norm). 
We present both $\delta_{\text{mean}}^{\text{high}(l)}$ (literal high-concrete usage, red line) and $\delta_{\text{mean}}^{\text{low}(l)}$ (figurative low-concrete usage, blue line) for every layer to visualize how their separation evolves across the layers.

\begin{figure}[t]
    \centering
    \includegraphics[width=0.99\linewidth]{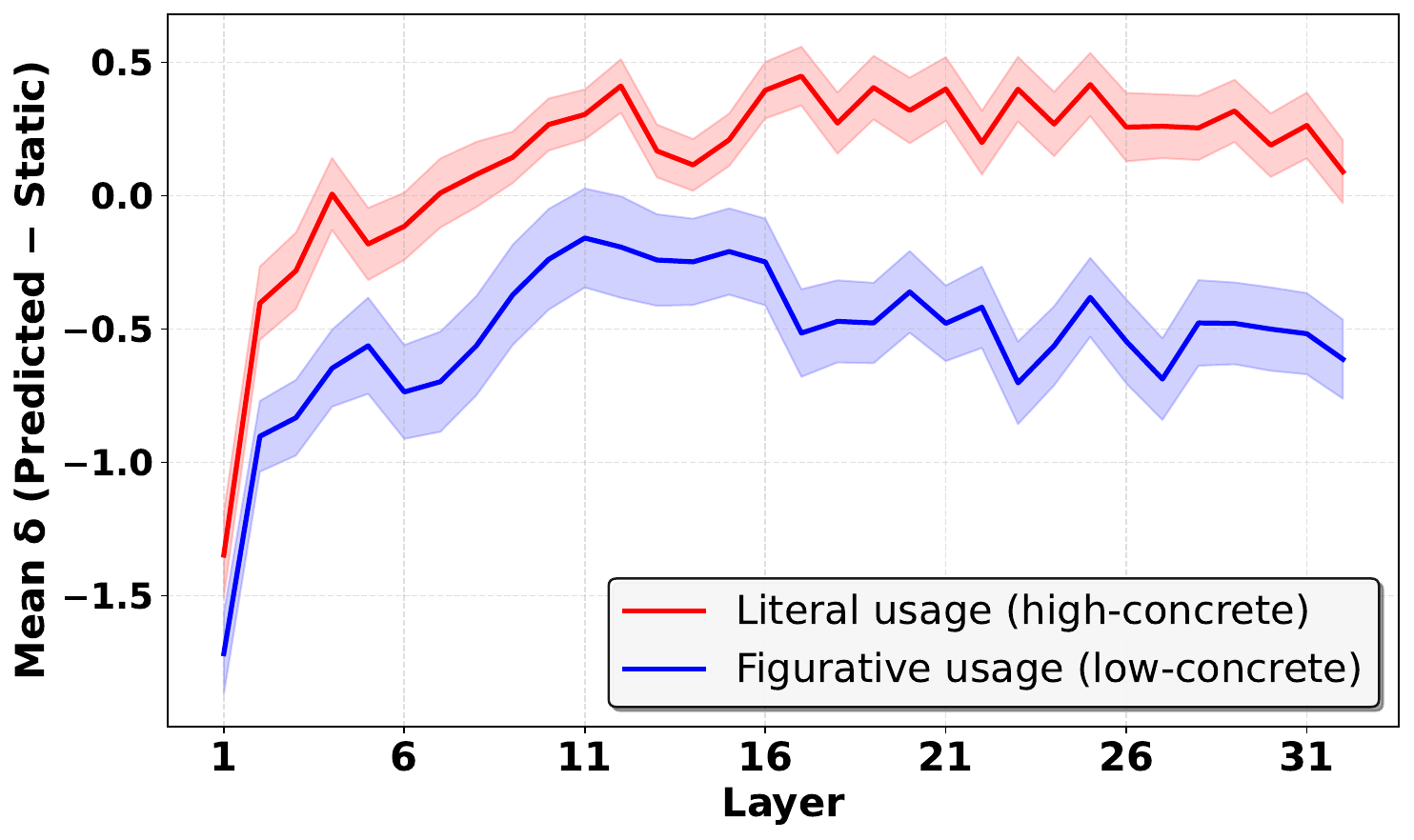}
    \caption{Mean $\delta$ across layers in Llama-3.1-8B, where $\delta = C_{\text{pred}} - C_{\text{static}}$. Positive values indicate higher predicted concreteness than the static norm (literal usage), while negative values indicate lower predicted concreteness (figurative usage). The early and consistent separation between high and low concrete curves shows that contextual shifts in concreteness are captured from early layers onward.}
    \label{fig:mean_delta}
\end{figure}

\noindent \textbf{Results.} Figure~\ref{fig:mean_delta} shows the results for Llama-3.1-8B; results for the remaining models are reported in Appendix~\ref{contextual_conc_rem}. Across all models, we observe a consistent pattern in which early layers already differentiate literal high-concrete from figurative low-concrete noun usage, with $\delta_{\text{mean}}^{\text{high}(l)}$ and $\delta_{\text{mean}}^{\text{low}(l)}$ separating from as early as layer~2 and remaining distinct through the final layer.

Importantly, the direction of the shift is systematic: literal high-concrete usages yield positive $\delta_{\text{mean}}^{\text{high}(l)}$, while figurative low-concrete usages yield negative $\delta_{\text{mean}}^{\text{low}(l)}$, indicating that the model adjusts concreteness estimates in the appropriate direction based on context. Appendix~\ref{app_verb_conc} shows the same experiment on verbs yields a lower separation.

\noindent \textbf{Takeaway.} LLMs encode concreteness and differentiate literal (high-concrete) and figurative (low-concrete) usage of a noun in the \textit{early layers}.

\section{Concreteness Geometric Subspace}

In this section, we examine how LLMs internally represent concreteness. We create a global concreteness axis and use it to classify figurative text.

\subsection{Creating Global Subspace}
\label{diffmean_svd}

We now want to characterize the geometric structure of concreteness representations by deriving a global subspace. For this, we first construct \textit{difference-in-means} (DiffMean) vector~\citep{diffmean} that captures the dominant direction along which concreteness varies in the model’s hidden space. DiffMean is a lightweight linear method that identifies directions associated with behavioral distinctions, offering mathematical simplicity along with direct interpretability and strong empirical performance~\citep{vennemeyer2025sycophancythingcausalseparation, yao2026rhetoricalquestionsllmrepresentations}.

\noindent \textbf{Constructing DiffMean.} We use the 25,000 Wikipedia sentences to create the DiffMean vector. Since DiffMean is the difference between the samples from positive and negative classes, we set two thresholds to gather high and low concrete sentences. Using static concreteness ratings (1--5) from \citet{concreteness}, we classify nouns with scores $\ge 4$ as high-concrete ($\mathcal{D}_{\text{high}}$) and scores $\le 2$ as low-concrete ($\mathcal{D}_{\text{low}}$). This yields a balanced sample of 2,256 high-concrete and 2,116 low-concrete sentence instances. Following our prompt-based probing setup, we extract the hidden representation of the last token that encodes concreteness information at every transformer layer~$l$ for both high and low concrete sets. For layer~$l$, we compute the mean representation of the last token for the high- and low-concrete groups:
\[
\mu_{\text{high}}^{(l)} =
\mathbb{E}_{i \sim \mathcal{D}_{\text{high}}}
\left[ \mathbf{h}^{(l)}_i \right],\quad
\mu_{\text{low}}^{(l)} =
\mathbb{E}_{i \sim \mathcal{D}_{\text{low}}}
\left[ \mathbf{h}^{(l)}_i \right],
\]
where $\mathbf{h}^{(l)}_i$ denotes the layer-$l$ hidden representation of the last token for sentence instance~$i$.
We define the \textit{DiffMean} at layer~$l$ as:
\[
\mathbf{w}^{(l)} = \mu_{\text{high}}^{(l)} - \mu_{\text{low}}^{(l)}.
\]

\noindent \textbf{Global Concreteness Subspace via SVD.} We seek to identify a compact set of linear directions that differentiate high and low concrete noun usage across the model’s layers. To do this, we first stack all the DiffMean vectors from all the layers, creating a global matrix $W$. We then perform singular value decomposition (SVD) on this matrix. This yields a set of orthogonal directions in the hidden state space ($V^\top$), ranked by how strongly they separate high and low concrete usages. We select the top-$k$ right singular vectors to form a global basis:
\[
B_k = V_{1:k}^\top,
\]
where $B_k$ contains the $k$ strongest directions along which high and low concrete usages differ. $B_k$ defines a layer-independent subspace capturing the dominant geometry of concreteness. By selecting $k$ in a single direction $(k=1)$, we test whether concreteness is encoded in a highly compressed, unidirectional space across layers of the LLM.

\subsection{Layer-wise Projection \& Results}
To evaluate the effectiveness of this global subspace, we use our prompt-based probing technique on the synthetic dataset and obtain, for each sentence, the hidden state $h^{(l)}$ at layer~$l$ that carries contextual concreteness information. We then project these embeddings into the global subspace:
\[
s(h^{(l)}) = B_k h^{(l)},
\]
where $B_k$ is the global concreteness subspace and $s(h^{(l)})$ are the $k$-dimensional projection scores. We then take the average of this scalar score across all dimensions in the model. Using these scores for high and low concrete sentences as positive and negative examples, we compute the area under the ROC curve (AUROC) at each layer, which quantifies how well the global subspace separates concrete usages across model depth.

\begin{figure}[t]
    \centering
    \includegraphics[width=0.99\linewidth]{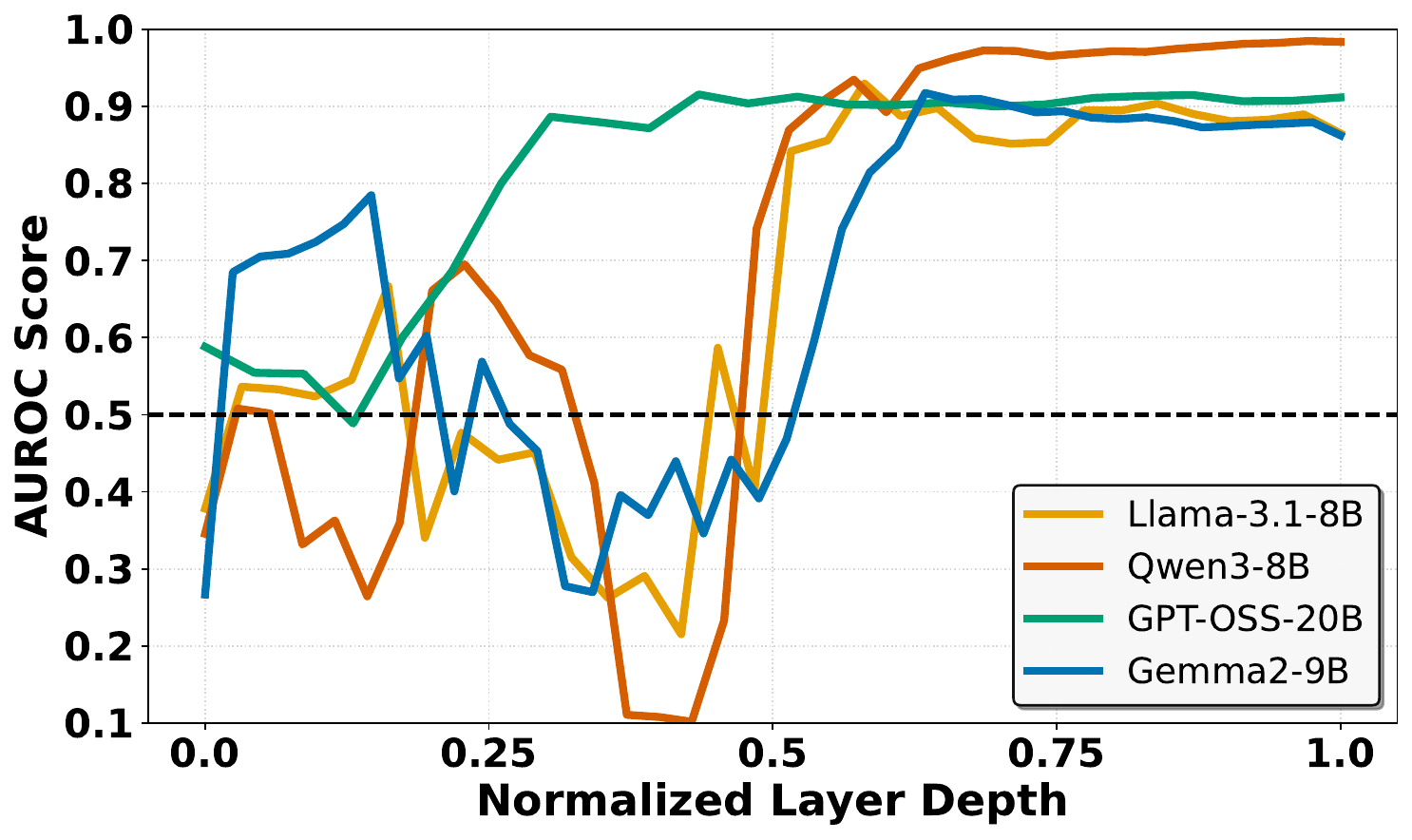}
    \caption{Layer-wise AUROC for separating high and low concrete noun usages using a global unidirectional concreteness axis across four LLMs, showing increasing separability and compression in middle and later layers.}
    \label{fig:subspace}
\end{figure}

\noindent\textbf{Results.} Figure~\ref{fig:subspace} reports AUROC scores for a unidirectional concreteness subspace ($k=1$) across all models with normalized layer depth. Individual model graphs and different $k$ values are provided in Appendix~\ref{appendix_subspace}. The results show that \textit{all models} compress concreteness into a \textbf{single global direction} in the later layers. The compression begins primarily in the middle layers. GPT-OSS-20B shows slightly different behavior as the compression begins in the early layers---we attribute this deviation due to its mixture-of-expert architecture~\citep{lo2025closerlookmixtureofexpertslarge}. For all models, AUROC score reach around $0.90$ from the middle layers through to the final layer, indicating that a unidirectional linear subspace is sufficient to distinguish concrete from abstract usage at those stages. 

\subsection{Figurative Text Classification}

Prior work on figurative text classification typically relies on training neural networks using the contextualized embeddings. Motivated by our finding that concreteness is encoded in a compressed one-directional subspace in the LLMs, we explore classifying figurative text using only this unidirectional concreteness axis 
\textit{without training}, and successfully achieve comparable performance to full-dimensional supervised training.

\noindent \textbf{Datasets.} We use three forms of figurative text and two datasets from each: i) Idioms - EPIE~\citep{saxena2020epiedatasetcorpuspossible} and MAGPIE~\citep{haagsma-etal-2020-magpie}, ii) Metaphor - VUA~\citep{vua} and MUNCH~\citep{munch}, iii) Metonymy - ConMeC~\citep{ghosh-jiang-2025-conmec} and MetFuse~\citep{ghosh2026metfusefigurativefusionmetonymy}. We select samples where the figurative expression is conveyed through the noun. The data preparing, preprocessing and distribution is provided in Appendix~\ref{figurative_text_dataset}.

\noindent \textbf{Methodology.}
Similar to the previous experiment, we use the 25,000 Wikipedia sentences to construct the DiffMean vector and get the one-directional concreteness subspace ($k=1$) using SVD. We then evaluate this geometric subspace on different figurative text classification datasets. For each dataset, we select the hidden representation of the target noun from a layer where our analysis showed strong single directional separability (layer 20 for Llama-3.1-8B). We then project the representation onto the learned unidirectional axis to obtain a single scalar projection score and compute the AUROC score. As a comparison, we also train a regression classifier with the same parameters as in Appendix~\ref{token_generation_prompt} on the last layer's \textit{full hidden state} (4,096-dimensional for Llama-3.1-8B) using 80-20 train-test split, averaged over 5 runs. This baseline represents the performance without dimensionality constraints. This setup allows us to directly measure how much discriminative signal for concreteness is preserved in the one-directional subspace.

\begin{table}[t]
\centering
\resizebox{0.98\linewidth}{!}{
\begin{tabular}{llcc}
\toprule
 &  & \multicolumn{2}{c}{\textbf{AUROC Score}} \\
\cmidrule(lr){3-4}
\textbf{Task} & \textbf{Dataset} 
& \begin{tabular}[c]{@{}c@{}}\textbf{Subspace}\\[-1pt]\textbf{(Zero-shot)}\end{tabular}
& \begin{tabular}[c]{@{}c@{}}\textbf{Full Rep.}\\[-1pt]\textbf{(Trained)}\end{tabular} \\
\midrule
\multirow{2}{*}{Idioms} 
    & MAGPIE & 95.2 & 98.5  \\
    &  EPIE & 95.3 & 99.2 \\
\midrule
\multirow{2}{*}{Metaphor} 
    & VUA & 95.7 & 97.6 \\
    & MUNCH & 93.2 & 95.1  \\
    
\midrule
\multirow{2}{*}{Metonymy} 
    & ConMeC & 60.2  & 62.6 \\
    & MetFuse &  85.7  & 96.3  \\
\bottomrule
\end{tabular}
}
\caption{AUROC scores for figurative text classification with Llama-3.1-8B, comparing a unidirectional concreteness subspace learned from Wikipedia and applied to downstream datasets against a full-representation classifier trained separately on each dataset.}

\label{table:figurative_f1}
\end{table}

\begingroup
\renewcommand\baselinestretch{0.99}
\begin{table*}[t]
    \NiceMatrixOptions {
    custom-line = {
       command = dashedmidrule ,
       tikz = { dashed } ,
       total-width = \pgflinewidth + \aboverulesep + \belowrulesep ,
     } }
    \centering
    \small
    \resizebox{\textwidth}{!}{
    \begin{NiceTabular}{|p{0.33\linewidth}|p{0.33\linewidth}|p{0.33\linewidth}|}
    \toprule
    \multicolumn{1}{c}{\bf Original Literal Sentence} & \multicolumn{1}{c}{\bf Unsteered Rewrite} & \multicolumn{1}{c}{\bf Figuratively Steered Rewrite} \\
    \midrule
    She struggled to push the heavy door open. & She struggled to force the heavy door open. & The heavy door battled her attempts to push it open.  \\ 
    \midrule
    The dirt path wound through the forest, the hikers followed it to a hidden waterfall. & The dirt path wound through the forest, leading the hikers to a hidden waterfall.  & A winding dirt path marched the hikers through the forest to a hidden waterfall.  \\ 
    \midrule
    The electoral results were to be announced later that day. & The electoral results were scheduled to be announced later that day. & Suspense engulfed the city for the electoral results later that day. \\ 
    \midrule
    The factory was shut down in 1972. & The factory closed in 1972. & The factory closed its door in 1972. \\ 
    \bottomrule
    
    \end{NiceTabular}
    }
    \caption{Example of concreteness-guided steering for rewriting literal sentences with Llama-3.1-8B. The original sentence and all the unsteered rewrites are literal. The steered rewrites are figurative.}
    \label{tab:lit-to-fig}
\end{table*}
\endgroup

\noindent\textbf{Results.} Table~\ref{table:figurative_f1} reports AUROC scores when using the concreteness subspace projection score to classify, compared to a classifier trained on the full 4,096-dimensional hidden state of Llama-3.1-8B. Results for other models are provided in Appendix~\ref{fig_text_rem}. All four models show consistent results. Across all figurative datasets, our subspace projection consistently recovers more than 95\% of the performance of the full classifier, despite it being a scalar projection on a unidirectional axis compared to the its original 4,096 dimensional embeddings. The high AUROC for metaphor and idiom datasets using the concreteness axis aligns with previous linguistic studies which states that concreteness of a term alters highly in metaphors and idioms~\citep{lakoff1993contemporary, glucksberg2001understanding}. In comparison, metonymy keeps the interpretation tied to a concrete entity or situation and the concreteness of a term shifts less~\citep{barcelona123}. This results in lower AUROC for metonymy compared to metaphors and idioms. MetFuse gets a relatively better score compared to ConMeC as it solely focuses on location metonymy, narrowing its scope. The gap between the unidirectional concreteness subspace with the fully trained classifier is more, tying to previous linguistic work regarding limited concreteness shift in metonymic interpretation of a term~\citep{panther1999metonymy}. These results indicate that the low-dimensional concreteness subspace can be used for figurative text identification.

\noindent \textbf{Takeaway.} LLMs \textit{compress concreteness into a unified one-directional subspace in the late layers}, and this compression \textit{emerges in the middle layers}. This subspace can be used for figurative text classification with significantly low latency and almost equal performance as a fully trained classifier. 

\section{Concreteness Controlled Steering}

The results above establish that LLMs encode literal-figurative variation along a highly compressible one-directional subspace in its deeper layers. We now assess whether this latent direction is merely representational or whether it plays a causal role in generative behavior. If this semantic dimension truly governs how the model interprets a noun in context, then directly manipulating hidden states along this axis should influence the figurativity of the generated text. We turn the concreteness subspace into a controllable knob, enabling post-hoc control over how literal or figurative the model's generations are. 

\noindent\textbf{Methodology.} We use the previous method to derive the unidirectional concreteness subspace using SVD over hidden representations of 25,000 Wikipedia sentences. This learned direction is subsequently applied to multiple figurative language datasets, enabling us to evaluate the cross-dataset transferability of the learned concreteness representation. This gives a unit direction $u$ along which concreteness varies in the model’s latent space. We then provide a sentence to the model and use a prompt to rewrite the input:  ``\textit{Rewrite the following sentence clearly and naturally:}''. As noted, the prompt does not contain any instructions about figurativeness or concreteness of the rewrite. During decoding, we intervene on the model’s hidden state $h^{(\ell)}$ at a selected layer $\ell$---chosen from layers where the one-directional subspace reliably encodes concreteness information (Figure~\ref{fig:subspace}). At this layer, we modify the hidden representation by adding an offset $\alpha$ along the concreteness axis:
\[
h_{\text{steer}}^{(\ell)} \;=\; h^{(\ell)} \;+\; \alpha \cdot u,
\]
where $\alpha$ controls the strength and direction of the intervention. For a more \textit{literal} rewrite, we steer in the direction of high concrete representations ($\alpha > 0$), and for a more \textit{figurative} rewrite, we steer in the opposite direction ($\alpha < 0$). The perturbed hidden state is then propagated through subsequent layers to produce the rewritten continuation.

To evaluate steering effectiveness, we use the synthetic dataset. We sample 100 figurative (low concrete) sentences when steering toward literal rewrites, and 100 literal (high concrete) sentences when steering toward figurative rewrites. For each sentence, we generate both steered ($\alpha=\pm 40$) and unsteered rewrites. In both cases, we use the same rewriting prompt without any hint regarding the intended expression of the rewrite. Then two independent human annotators rate whether the output exhibits the intended literal or figurative expression. We intervene at layer 20 for Llama-3.1-8B, 25 for Qwen3-8B, 27 for Gemma2-9B and 15 for GPT-OSS-20B.

\begin{table}[t]
\centering
\resizebox{0.98\linewidth}{!}{
\begin{tabular}{lcccc}
\toprule
 & \multicolumn{2}{c}{\textbf{Lit $\rightarrow$ Fig}} & \multicolumn{2}{c}{\textbf{Fig $\rightarrow$ Lit}} \\
\cmidrule(lr){2-3} \cmidrule(lr){4-5}
\textbf{Model} & \textbf{Steer} & \textbf{Unsteer} &  \textbf{Steer} & \textbf{Unsteer} \\
\midrule
Llama-3.1-8B & \textbf{12} & 0 & \textbf{71} & 42 \\
Qwen3-8B & \textbf{10} & 0 & \textbf{68} & 45 \\
Gemma2-9B & \textbf{9} & 0 & \textbf{67} & 52 \\
GPT-OSS-20B & \textbf{15} & 0 & \textbf{75} & 39 \\
\bottomrule
\end{tabular}
}
\caption{Number of sentences judged by humans to be successfully steered in a sample of 100. \textit{Lit $\rightarrow$ Fig} indicates original sentence was literal, the LLM rewrite is figurative. \textit{Fig $\rightarrow$ Lit} is when input is figurative, rewrite is literal.}
\label{table:steering}
\end{table}

\noindent\textbf{Results.} Table~\ref{tab:lit-to-fig} shows some qualitative examples of unsteered vs steered rewrites going from literal to figurative. Table~\ref{table:steering} shows the quantitative result of the steering-based rewriting experiment. We used human annotators to evaluate 100 sentences under both figurative and literal steering conditions, as well as their unsteered counterparts. For literal inputs, all models produce literal paraphrases in the unsteered condition. Steering towards the low-concrete direction ($\alpha=-40$) induces figurative reinterpretations for around 10\% of the sentences. For figurative inputs, unsteered rewriting already produces literal paraphrases for 40--50\% of the cases, reflecting a strong literal bias in LLMs~\citep{chakrabarty-etal-2022-rocket}. Our concreteness-based steering consistently improves this figure to around 70\%, showing that shifting hidden states along the identified concreteness axis makes the model more likely to rewrite the figurative sentence in a literal sense. Table~\ref{tab:fig-to-lit} in Appendix~\ref{app_steering_example} shows qualitative examples of steering from figurative to literal. 

The results reveal that steering from figurative to literal usage is more effective than steering from literal to figurative usage. This aligns with the observation that transforming literal text to figurative is difficult due to semantic and meaning abstraction challenges~\citep{stowe-etal-2021-metaphor, chakrabarty-etal-2021-mermaid}. Figurative language requires controlled conceptual mapping while preserving meaning, which makes generation harder than literal instances~\citep{lai_survey}.

Nevertheless, the ability to induce even moderate figurative reinterpretation from literal inputs without modifying model parameters or providing a figurativity prompt suggests that the learned concreteness direction is a \textit{causally relevant} control dimension. We believe that this result opens opportunities for future work on controllable figurative generation, including designing steering vectors derived from richer figurativity phenomena (e.g., metonymy vs.\ metaphor distinctions) and combining causal steering with explicit task prompts to further enhance stylistic expressivity.

\noindent \textbf{Takeaway.} The one-directional geometric axis of concreteness could be used as a causally manipulable handle for modifying the figurative-literal interpretation of text, enabling controlled rewriting without model retraining or task-specific prompts.

\section{Conclusion}

Our work provides a comprehensive analysis of how LLMs represent and process concreteness internally. We show that the early layers encode concreteness, and differentiate high-concrete (literal) and low-concrete (figurative) usage of a term. In later layers, concreteness is compressed into a shared direction in representation space, and the compression emerges from the middle layers. This structure generalizes across models and supports zero-shot figurative language classification with significantly low latency and close to a fully trained classifier. Finally, we demonstrate that manipulating this concreteness direction causally steers generation between literal and figurative interpretations.

\section*{Limitations}

While our work analyzes the internals of different LLMs in understanding concreteness and shows promising future research direction, certain limitations exist. First, our study does not use human-annotated judgments for contextual concreteness. We use a trained regression model and concreteness information carrying token to predict the concreteness score. The predicted scores show very high correlation with the human annotated scores ($\approx$0.98). However, incorporating human annotations would provide a stronger grounding for evaluating contextual variation and remains an important direction for future work.

We construct a unidirectional concreteness subspace using DiffMean vectors derived from high and low concrete noun usages. This subspace yields competitive performance for figurative language classification and enables controllable rewriting between figurative and literal expressions. However, it may be possible that this subspace does not isolate concreteness in a strictly independent manner. The identified direction may also encode other correlated semantic or linguistic factors, and intervening along this axis does not guarantee that only concreteness-related properties are affected. Identifying overlapping factors in a geometric subspace and disentangling them remains challenging. We view this as an interesting and important direction and leave it for future work.

While concreteness is strongly related to figurative language, the two are not equivalent. A reduction in contextual concreteness does not uniquely imply figurative usage. For instance, word sense disambiguation phenomena can also involve shifts from highly concrete nouns to less concrete contextual usage without invoking figurative language (\textit{root} of a tree vs. square \textit{root} of a number). To mitigate this confound, we explicitly identify and remove such cases when constructing our synthetic dataset, ensuring that concreteness shifts are primarily driven by figurative usage. Consequently, our experiments evaluate the concreteness axis in a controlled setting where concreteness-based distinctions align with figurative contrasts. However, this does not imply that all figurative language can be reduced to concreteness differences, nor that concreteness alone is sufficient to fully characterize figurative meaning.

\section*{Ethical Considerations}

The natural language data used in our experiments originates from Wikipedia, which may contain historical and demographic biases; however, our study focuses exclusively on representational differences in word concreteness and does not involve any applications with social impact. Synthetic evaluation examples were generated using a commercial LLM (GPT-5.1), and human annotation was limited to verifying concreteness usage rather than sensitive attributes.

\section*{Acknowledgments}

We thank the CincyNLP group for their suggestions and feedback. We also thank the anonymous ACL reviewers for
their insightful suggestions.

\bibliography{main}

\begin{thebibliography}{59}
\providecommand{\natexlab}[1]{#1}

\bibitem[{Barcelona(2003)}]{barcelona123}
Antonio Barcelona. 2003.
\newblock \emph{Metaphor and Metonymy at the Crossroads: A Cognitive Perspective}.
\newblock Mouton de Gruyter.

\bibitem[{Barsalou(1999)}]{Barsalou_1999}
Lawrence~W Barsalou. 1999.
\newblock \href {https://doi.org/10.1017/S0140525X99002149} {Perceptual symbol systems}.
\newblock \emph{Behavioral and Brain Sciences}, 22(4):577--660.

\bibitem[{Beigman~Klebanov et~al.(2015)Beigman~Klebanov, Leong, and Flor}]{beigman-klebanov-etal-2015-supervised}
Beata Beigman~Klebanov, Chee~Wee Leong, and Michael Flor. 2015.
\newblock \href {https://doi.org/10.3115/v1/W15-1402} {Supervised word-level metaphor detection: Experiments with concreteness and reweighting of examples}.
\newblock In \emph{Proceedings of the Third Workshop on Metaphor in {NLP}}, pages 11--20, Denver, Colorado. Association for Computational Linguistics.

\bibitem[{Bonin et~al.(2018)Bonin, M{\'e}ot, and Bugaiska}]{Bonin2018FrenchConcreteness}
P.~Bonin, A.~M{\'e}ot, and A.~Bugaiska. 2018.
\newblock \href {https://doi.org/10.3758/s13428-018-1014-y} {Concreteness norms for 1,659 french words: Relationships with other psycholinguistic variables and word recognition times}.
\newblock \emph{Behavior Research Methods}, 50(6):2366--2387.

\bibitem[{Brown et~al.(2020)Brown, Mann, Ryder, Subbiah, Kaplan, Dhariwal, Neelakantan, Shyam, Sastry, Askell et~al.}]{brown2020languagemodelsfewshotlearners}
Tom~B. Brown, Benjamin Mann, Nick Ryder, Melanie Subbiah, Jared Kaplan, Prafulla Dhariwal, Arvind Neelakantan, Pranav Shyam, Girish Sastry, Amanda Askell, and 1 others. 2020.
\newblock \href {https://arxiv.org/abs/2005.14165} {Language models are few-shot learners}.
\newblock \emph{Preprint}, arXiv:2005.14165.

\bibitem[{Brysbaert et~al.(2014)Brysbaert, Warriner, and Kuperman}]{concreteness}
Marc Brysbaert, Amy~Beth Warriner, and Victor Kuperman. 2014.
\newblock \href {https://doi.org/10.3758/s13428-013-0403-5} {Concreteness ratings for 40 thousand generally known english word lemmas}.
\newblock \emph{Behavior Research Methods}, 46(3):904--911.

\bibitem[{Cacciari and Tabossi(1993)}]{cacciari1993idioms}
Cristina Cacciari and Patrizia Tabossi, editors. 1993.
\newblock \emph{Idioms: Processing, Structure, and Interpretation}.
\newblock Lawrence Erlbaum Associates, Hillsdale, NJ, USA.

\bibitem[{Chakrabarty et~al.(2022)Chakrabarty, Choi, and Shwartz}]{chakrabarty-etal-2022-rocket}
Tuhin Chakrabarty, Yejin Choi, and Vered Shwartz. 2022.
\newblock \href {https://doi.org/10.1162/tacl_a_00478} {It{'}s not rocket science: Interpreting figurative language in narratives}.
\newblock \emph{Transactions of the Association for Computational Linguistics}, 10:589--606.

\bibitem[{Chakrabarty et~al.(2021)Chakrabarty, Zhang, Muresan, and Peng}]{chakrabarty-etal-2021-mermaid}
Tuhin Chakrabarty, Xurui Zhang, Smaranda Muresan, and Nanyun Peng. 2021.
\newblock \href {https://doi.org/10.18653/v1/2021.naacl-main.336} {{MERMAID}: Metaphor generation with symbolism and discriminative decoding}.
\newblock In \emph{Proceedings of the 2021 Conference of the North American Chapter of the Association for Computational Linguistics (NAACL 2021)}.

\bibitem[{Charbonnier and Wartena(2019)}]{charbonnier-wartena-2019-predicting}
Jean Charbonnier and Christian Wartena. 2019.
\newblock \href {https://doi.org/10.18653/v1/W19-0415} {Predicting word concreteness and imagery}.
\newblock In \emph{Proceedings of the 13th International Conference on Computational Semantics - Long Papers}, pages 176--187, Gothenburg, Sweden. Association for Computational Linguistics.

\bibitem[{Devlin et~al.(2019)Devlin, Chang, Lee, and Toutanova}]{devlin-etal-2019-bert}
Jacob Devlin, Ming-Wei Chang, Kenton Lee, and Kristina Toutanova. 2019.
\newblock \href {https://doi.org/10.18653/v1/N19-1423} {{BERT}: Pre-training of deep bidirectional transformers for language understanding}.
\newblock In \emph{Proceedings of the 2019 Conference of the North {A}merican Chapter of the Association for Computational Linguistics (NAACL 2019)}.

\bibitem[{Frassinelli et~al.(2017)Frassinelli, Naumann, Utt, and Schulte~m Walde}]{frassinelli-etal-2017-contextual}
Diego Frassinelli, Daniela Naumann, Jason Utt, and Sabine Schulte~m Walde. 2017.
\newblock \href {https://aclanthology.org/W17-6910/} {Contextual characteristics of concrete and abstract words}.
\newblock In \emph{Proceedings of the 12th International Conference on Computational Semantics ({IWCS}) {---} Short papers}.

\bibitem[{{Gemma Team} et~al.(2024){Gemma Team}, Riviere, Pathak, Sessa, Hardin, Bhupatiraju, Hussenot, Mesnard, Shahriari, Ramé, Ferret, Liu, Tafti, Friesen, Casbon, Ramos, Kumar, Lan, Jerome, Tsitsulin, Vieillard, Stanczyk, Girgin, Momchev, Hoffman, Thakoor, Grill, Neyshabur, Bachem, Walton, Severyn, Parrish, Ahmad, Hutchison, Abdagic, Carl, Shen, Brock, Coenen, Laforge, Paterson, Bastian, Piot, Wu, Royal, Chen, Kumar, Perry, Welty, Choquette-Choo, Sinopalnikov, Weinberger, Vijaykumar, Rogozińska, Herbison, Bandy, Wang, Noland, Moreira, Senter, Eltyshev, Visin, Rasskin, Wei, Cameron, Martins, Hashemi, Klimczak-Plucińska, Batra, Dhand, Nardini, Mein, Zhou, Svensson, Stanway, Chan, Zhou, Carrasqueira, Iljazi, Becker, Fernandez, van Amersfoort, Gordon, Lipschultz, Newlan, yeong Ji, Mohamed, Badola, Black, Millican, McDonell, Nguyen, Sodhia, Greene, Sjoesund, Usui, Sifre, Heuermann, Lago, McNealus, Soares, Kilpatrick, Dixon, Martins, Reid, Singh, Iverson, Görner, Velloso, Wirth, Davidow, Miller, Rahtz,
  Watson, Risdal, Kazemi, Moynihan, Zhang, Kahng, Park, Rahman, Khatwani, Dao, Bardoliwalla, Devanathan, Dumai, Chauhan, Wahltinez, Botarda, Barnes, Barham, Michel, Jin, Georgiev, Culliton, Kuppala, Comanescu, Merhej, Jana, Rokni, Agarwal, Mullins, Saadat, Carthy, Cogan, Perrin, Arnold, Krause, Dai, Garg, Sheth, Ronstrom, Chan, Jordan, Yu, Eccles, Hennigan, Kocisky, Doshi, Jain, Yadav, Meshram, Dharmadhikari, Barkley, Wei, Ye, Han, Kwon, Xu, Shen, Gong, Wei, Cotruta, Kirk, Rao, Giang, Peran, Warkentin, Collins, Barral, Ghahramani, Hadsell, Sculley, Banks, Dragan, Petrov, Vinyals, Dean, Hassabis, Kavukcuoglu, Farabet, Buchatskaya, Borgeaud, Fiedel, Joulin, Kenealy, Dadashi, and Andreev}]{gemma_2024}
{Gemma Team}, Morgane Riviere, Shreya Pathak, Pier~Giuseppe Sessa, Cassidy Hardin, Surya Bhupatiraju, Léonard Hussenot, Thomas Mesnard, Bobak Shahriari, Alexandre Ramé, Johan Ferret, Peter Liu, Pouya Tafti, Abe Friesen, Michelle Casbon, Sabela Ramos, Ravin Kumar, Charline~Le Lan, Sammy Jerome, and 179 others. 2024.
\newblock \href {https://arxiv.org/abs/2408.00118} {Gemma 2: Improving open language models at a practical size}.
\newblock \emph{Preprint}, arXiv:2408.00118.

\bibitem[{Ghosh and Jiang(2025)}]{ghosh-jiang-2025-conmec}
Saptarshi Ghosh and Tianyu Jiang. 2025.
\newblock \href {https://doi.org/10.18653/v1/2025.naacl-long.330} {{C}on{M}e{C}: A dataset for metonymy resolution with common nouns}.
\newblock In \emph{Proceedings of the 2025 Conference of the Nations of the Americas Chapter of the Association for Computational Linguistics (NAACL 2025)}.

\bibitem[{Ghosh and Jiang(2026)}]{ghosh2026metfusefigurativefusionmetonymy}
Saptarshi Ghosh and Tianyu Jiang. 2026.
\newblock \href {https://arxiv.org/abs/2604.12919} {Metfuse: Figurative fusion between metonymy and metaphor}.
\newblock \emph{Preprint}, arXiv:2604.12919.

\bibitem[{Ghosh et~al.(2026)Ghosh, Liu, and Jiang}]{ghosh-etal-2026-computational}
Saptarshi Ghosh, Linfeng Liu, and Tianyu Jiang. 2026.
\newblock \href {https://doi.org/10.18653/v1/2026.eacl-long.92} {A computational approach to visual metonymy}.
\newblock In \emph{Proceedings of the 19th Conference of the {E}uropean Chapter of the {A}ssociation for {C}omputational {L}inguistics (EACL 2026)}.

\bibitem[{Glucksberg(2001)}]{glucksberg2001understanding}
Sam Glucksberg. 2001.
\newblock \href {https://academic.oup.com/book/32733} {\emph{Understanding Figurative Language: From Metaphor to Idioms}}.
\newblock Oxford University Press, New York, NY, USA.

\bibitem[{Grattafiori et~al.(2024)Grattafiori, Dubey, Jauhri, Pandey, Kadian, Al-Dahle, Letman, Mathur, Schelten, Vaughan, Yang, Fan, and others.}]{grattafiori2024llama3herdmodels}
Aaron Grattafiori, Abhimanyu Dubey, Abhinav Jauhri, Abhinav Pandey, Abhishek Kadian, Ahmad Al-Dahle, Aiesha Letman, Akhil Mathur, Alan Schelten, Alex Vaughan, Amy Yang, Angela Fan, and others. 2024.
\newblock \href {https://arxiv.org/abs/2407.21783} {The llama 3 herd of models}.
\newblock \emph{Preprint}, arXiv:2407.21783.

\bibitem[{Guasch et~al.(2016)Guasch, Ferr{\'e}, and Fraga}]{Guasch2016SpanishNorms}
M.~Guasch, P.~Ferr{\'e}, and I.~Fraga. 2016.
\newblock \href {https://doi.org/10.3758/s13428-015-0684-y} {Spanish norms for affective and lexico-semantic variables for 1,400 words}.
\newblock \emph{Behavior Research Methods}, 48(4):1358--1369.

\bibitem[{Haagsma et~al.(2020)Haagsma, Bos, and Nissim}]{haagsma-etal-2020-magpie}
Hessel Haagsma, Johan Bos, and Malvina Nissim. 2020.
\newblock \href {https://aclanthology.org/2020.lrec-1.35/} {{MAGPIE}: A large corpus of potentially idiomatic expressions}.
\newblock In \emph{Proceedings of the Twelfth Language Resources and Evaluation Conference}, pages 279--287, Marseille, France. European Language Resources Association.

\bibitem[{Hessel et~al.(2018)Hessel, Mimno, and Lee}]{hessel-etal-2018-quantifying}
Jack Hessel, David Mimno, and Lillian Lee. 2018.
\newblock \href {https://doi.org/10.18653/v1/N18-1199} {Quantifying the visual concreteness of words and topics in multimodal datasets}.
\newblock In \emph{Proceedings of the 2018 Conference of the North {A}merican Chapter of the Association for Computational Linguistics (NAACL 2018)}.

\bibitem[{Holyoak and Stamenkovi{\'c}(2018)}]{Holyoak2018MetaphorReview}
Keith~J. Holyoak and Du{\v{s}}an Stamenkovi{\'c}. 2018.
\newblock \href {https://doi.org/10.1037/bul0000145} {Metaphor comprehension: A critical review of theories and evidence}.
\newblock \emph{Psychological Bulletin}, 144(6):641--671.

\bibitem[{Iaia et~al.(2025)Iaia, Choksi, Wiebers, Roig, and Fiebach}]{iaia2025representationalalignmenthumanslanguage}
Cosimo Iaia, Bhavin Choksi, Emily Wiebers, Gemma Roig, and Christian~J. Fiebach. 2025.
\newblock \href {https://arxiv.org/abs/2505.15682} {The representational alignment between humans and language models is implicitly driven by a concreteness effect}.
\newblock \emph{Preprint}, arXiv:2505.15682.

\bibitem[{Jessen et~al.(2000)Jessen, Heun, Erb, Granath, Klose, Papassotiropoulos, and Grodd}]{JESSEN2000103}
F.~Jessen, R.~Heun, M.~Erb, D.-O. Granath, U.~Klose, A.~Papassotiropoulos, and W.~Grodd. 2000.
\newblock \href {https://doi.org/10.1006/brln.2000.2340} {The concreteness effect: Evidence for dual coding and context availability}.
\newblock \emph{Brain and Language}, 74(1):103--112.

\bibitem[{Kewenig et~al.(2025)Kewenig, Skipper, and Vigliocco}]{Kewenig2025AMT}
Viktor Kewenig, Jeremy~I. Skipper, and Gabriella Vigliocco. 2025.
\newblock \href {https://api.semanticscholar.org/CorpusID:279984017} {A multimodal transformer-based tool for automatic generation of concreteness ratings across languages}.
\newblock \emph{Communications Psychology}, 3.

\bibitem[{Lai and Nissim(2024)}]{lai_survey}
Huiyuan Lai and Malvina Nissim. 2024.
\newblock \href {https://doi.org/10.1145/3654795} {A survey on automatic generation of figurative language: From rule-based systems to large language models}.
\newblock \emph{ACM Comput. Surv.}, 56(10).

\bibitem[{Lai et~al.(2019)Lai, Howerton, and Desai}]{Lai2019ActionMetaphorERP}
Vicky~T. Lai, Odessa Howerton, and Rutvik~H. Desai. 2019.
\newblock \href {https://doi.org/10.1016/j.brainres.2019.03.005} {Concrete processing of action metaphors: Evidence from erp}.
\newblock \emph{Brain Research}, 1714:202--209.

\bibitem[{Lakoff(1993)}]{lakoff1993contemporary}
George Lakoff. 1993.
\newblock \href {https://doi.org/10.1017/CBO9781139173865.013} {The contemporary theory of metaphor}.
\newblock In Andrew Ortony, editor, \emph{Metaphor and Thought}, 2 edition, pages 202--251. Cambridge University Press, Cambridge, UK.

\bibitem[{Lakoff and Johnson(1980)}]{lakoff1980}
George Lakoff and Mark Johnson. 1980.
\newblock \emph{Metaphors We Live By}.
\newblock University of Chicago Press, Chicago.

\bibitem[{Lo et~al.(2025)Lo, Huang, Qiu, Wang, and Fu}]{lo2025closerlookmixtureofexpertslarge}
Ka~Man Lo, Zeyu Huang, Zihan Qiu, Zili Wang, and Jie Fu. 2025.
\newblock \href {https://doi.org/10.18653/v1/2025.findings-naacl.251} {A closer look into mixture-of-experts in large language models}.
\newblock In \emph{Findings of the Association for Computational Linguistics: NAACL 2025 (Findings of NAACL 2025)}.

\bibitem[{Marks and Tegmark(2024)}]{diffmean}
Samuel Marks and Max Tegmark. 2024.
\newblock \href {https://arxiv.org/abs/2310.06824} {The geometry of truth: Emergent linear structure in large language model representations of true/false datasets}.
\newblock \emph{Preprint}, arXiv:2310.06824.

\bibitem[{Maudslay et~al.(2020)Maudslay, Pimentel, Cotterell, and Teufel}]{hall-maudslay-etal-2020-metaphor}
Rowan~Hall Maudslay, Tiago Pimentel, Ryan Cotterell, and Simone Teufel. 2020.
\newblock \href {https://doi.org/10.18653/v1/2020.figlang-1.30} {Metaphor detection using context and concreteness}.
\newblock In \emph{Proceedings of the Second Workshop on Figurative Language Processing}, pages 221--226, Online. Association for Computational Linguistics.

\bibitem[{Mialon et~al.(2023)Mialon, Dessì, Lomeli, Nalmpantis, Pasunuru, Raileanu, Rozière, Schick, Dwivedi-Yu, Celikyilmaz, Grave, LeCun, and Scialom}]{mialon2023augmentedlanguagemodelssurvey}
Grégoire Mialon, Roberto Dessì, Maria Lomeli, Christoforos Nalmpantis, Ram Pasunuru, Roberta Raileanu, Baptiste Rozière, Timo Schick, Jane Dwivedi-Yu, Asli Celikyilmaz, Edouard Grave, Yann LeCun, and Thomas Scialom. 2023.
\newblock \href {https://arxiv.org/abs/2302.07842} {Augmented language models: a survey}.
\newblock \emph{Preprint}, arXiv:2302.07842.

\bibitem[{Mon et~al.(2021)Mon, Nencheva, Citron, Lew-Williams, and Goldberg}]{MON2021104285}
Serena~K. Mon, Mira Nencheva, Francesca~M.M. Citron, Casey Lew-Williams, and Adele~E. Goldberg. 2021.
\newblock \href {https://doi.org/10.1016/j.jml.2021.104285} {Conventional metaphors elicit greater real-time engagement than literal paraphrases or concrete sentences}.
\newblock \emph{Journal of Memory and Language}, 121:104285.

\bibitem[{Montefinese et~al.(2023)Montefinese, Gregori, Ravelli, Varvara, and Radicioni}]{Montefinese2023CONcreTEXT}
M.~Montefinese, L.~Gregori, A.~A. Ravelli, R.~Varvara, and D.~P. Radicioni. 2023.
\newblock \href {https://doi.org/10.1371/journal.pone.0293031} {Concretext norms: Concreteness ratings for italian and english words in context}.
\newblock \emph{PLOS ONE}, 18(10):e0293031.

\bibitem[{Montefinese et~al.(2025)Montefinese, Visalli, Angrilli, and Ambrosini}]{Montefinese2025ConcretenessERP}
M.~Montefinese, A.~Visalli, A.~Angrilli, and E.~Ambrosini. 2025.
\newblock \href {https://doi.org/10.1111/psyp.70074} {Fine-grained concreteness effects on word processing and representation across three tasks: An erp study}.
\newblock \emph{Psychophysiology}, 62(5):e70074.

\bibitem[{Muraki et~al.(2023)Muraki, Abdalla, Brysbaert, and Pexman}]{Muraki2022MWEConcreteness}
Emiko~J Muraki, Summer Abdalla, Marc Brysbaert, and Penny~M Pexman. 2023.
\newblock \href {https://doi.org/10.3758/s13428-022-01912-6} {Concreteness ratings for 62,000 english multiword expressions}.
\newblock \emph{Behavior research methods}, 55(5):2522--2531.

\bibitem[{OpenAI et~al.(2025)OpenAI, :, Agarwal, Ahmad, Ai, Altman, Applebaum, Arbus, Arora, Bai, Baker, Bao, Barak, Bennett, Bertao, and others.}]{openai2025gptoss120bgptoss20bmodel}
OpenAI, :, Sandhini Agarwal, Lama Ahmad, Jason Ai, Sam Altman, Andy Applebaum, Edwin Arbus, Rahul~K. Arora, Yu~Bai, Bowen Baker, Haiming Bao, Boaz Barak, Ally Bennett, Tyler Bertao, and others. 2025.
\newblock \href {https://arxiv.org/abs/2508.10925} {gpt-oss-120b gpt-oss-20b model card}.
\newblock \emph{Preprint}, arXiv:2508.10925.

\bibitem[{Panther and Radden(1999)}]{panther1999metonymy}
Klaus-Uwe Panther and G{\"u}nter Radden, editors. 1999.
\newblock \emph{Metonymy in Language and Thought}, volume~4 of \emph{Human Cognitive Processing}.
\newblock John Benjamins Publishing Company, Amsterdam / Philadelphia.

\bibitem[{Pollock(2018)}]{lewis}
Lewis Pollock. 2018.
\newblock Concepts and concreteness in psycholinguistics.

\bibitem[{Puccetti et~al.(2024)Puccetti, Collacciani, Ravelli, Esuli, and Bolognesi}]{puccetti-etal-2024-abricot}
Giovanni Puccetti, Claudia Collacciani, Andrea~Amelio Ravelli, Andrea Esuli, and Marianna Bolognesi. 2024.
\newblock \href {https://aclanthology.org/2024.clicit-1.128/} {{ABRICOT} - {AB}st{R}actness and inclusiveness in {CO}ntex{T}: A {CALAMITA} challenge}.
\newblock In \emph{Proceedings of the Tenth Italian Conference on Computational Linguistics (CLiC-it 2024)}, pages 1161--1167, Pisa, Italy. CEUR Workshop Proceedings.

\bibitem[{Saxena and Paul(2020)}]{saxena2020epiedatasetcorpuspossible}
Prateek Saxena and Soma Paul. 2020.
\newblock \href {https://arxiv.org/abs/2006.09479} {Epie dataset: A corpus for possible idiomatic expressions}.
\newblock \emph{Preprint}, arXiv:2006.09479.

\bibitem[{Schwanenflugel and Shoben(1983)}]{Schwanenflugel_Shoben_1983}
Paula~J. Schwanenflugel and Edward~J. Shoben. 1983.
\newblock \href {https://doi.org/10.1037/0278-7393.9.1.82} {Differential context effects in the comprehension of abstract and concrete verbal materials}.
\newblock \emph{Journal of Experimental Psychology: Learning, Memory, and Cognition}, 9(1):82–102.

\bibitem[{Steen et~al.(2010)Steen, Dorst, Herrmann, Kaal, Krennmayr, and Pasma}]{vua}
Gerard~J. Steen, Agnes~G. Dorst, J.~Berenike Herrmann, Anna~A. Kaal, Thomas Krennmayr, and Trijntje Pasma. 2010.
\newblock \href {https://benjamins.com/catalog/celcr.14?srsltid=AfmBOooFl3yxMvFzm_I1lskLJsUQ7RzOHK6s6zDvPrUKAnMs-putQaHd} {\emph{A Method for Linguistic Metaphor Identification: From MIP to MIPVU}}.
\newblock John Benjamins Publishing Company, Amsterdam.

\bibitem[{Stowe et~al.(2021)Stowe, Chakrabarty, Peng, Muresan, and Gurevych}]{stowe-etal-2021-metaphor}
Kevin Stowe, Tuhin Chakrabarty, Nanyun Peng, Smaranda Muresan, and Iryna Gurevych. 2021.
\newblock \href {https://doi.org/10.18653/v1/2021.acl-long.524} {Metaphor generation with conceptual mappings}.
\newblock In \emph{Proceedings of the 59th Annual Meeting of the Association for Computational Linguistics and the 11th International Joint Conference on Natural Language Processing (ACL 2021)}.

\bibitem[{Su et~al.(2021)Su, Chen, Fu, and Chen}]{SU2021166}
Chang Su, Weijie Chen, Ze~Fu, and Yijiang Chen. 2021.
\newblock \href {https://doi.org/10.1016/j.neucom.2020.11.051} {Multimodal metaphor detection based on distinguishing concreteness}.
\newblock \emph{Neurocomputing}, 429:166--173.

\bibitem[{Tater et~al.(2022)Tater, Frassinelli, and Schulte~im Walde}]{tater-etal-2022-concreteness}
Tarun Tater, Diego Frassinelli, and Sabine Schulte~im Walde. 2022.
\newblock \href {https://doi.org/10.18653/v1/2022.aacl-srw.13} {Concreteness vs. abstractness: A selectional preference perspective}.
\newblock In \emph{Proceedings of the 2nd Conference of the Asia-Pacific Chapter of the Association for Computational Linguistics and the 12th International Joint Conference on Natural Language Processing: Student Research Workshop}, pages 92--98, Online. Association for Computational Linguistics.

\bibitem[{Tong et~al.(2024)Tong, Choenni, Lewis, and Shutova}]{munch}
Xiaoyu Tong, Rochelle Choenni, Martha Lewis, and Ekaterina Shutova. 2024.
\newblock \href {https://doi.org/10.18653/v1/2024.acl-long.193} {Metaphor understanding challenge dataset for {LLM}s}.
\newblock In \emph{Proceedings of the 62nd Annual Meeting of the Association for Computational Linguistics (ACL 2024)}.

\bibitem[{Troche et~al.(2017)Troche, Crutch, and Reilly}]{troche2017}
Joshua Troche, Sebastian Crutch, and Jamie Reilly. 2017.
\newblock \href {https://doi.org/10.3389/fpsyg.2017.01787} {Defining a conceptual topography of word concreteness: Clustering properties of emotion, sensation, and magnitude among 750 english words}.
\newblock \emph{Frontiers in Psychology}, 8:1787.

\bibitem[{Tsvetkov et~al.(2014)Tsvetkov, Boytsov, Gershman, Nyberg, and Dyer}]{tsvetkov-etal-2014-metaphor}
Yulia Tsvetkov, Leonid Boytsov, Anatole Gershman, Eric Nyberg, and Chris Dyer. 2014.
\newblock \href {https://doi.org/10.3115/v1/P14-1024} {Metaphor detection with cross-lingual model transfer}.
\newblock In \emph{Proceedings of the 52nd Annual Meeting of the Association for Computational Linguistics (ACL 2014)}.

\bibitem[{Tsvetkov et~al.(2013)Tsvetkov, Mukomel, and Gershman}]{tsvetkov-etal-2013-cross}
Yulia Tsvetkov, Elena Mukomel, and Anatole Gershman. 2013.
\newblock \href {https://aclanthology.org/W13-0906/} {Cross-lingual metaphor detection using common semantic features}.
\newblock In \emph{Proceedings of the First Workshop on Metaphor in {NLP}}, pages 45--51, Atlanta, Georgia. Association for Computational Linguistics.

\bibitem[{Vennemeyer et~al.(2025)Vennemeyer, Duong, Zhan, and Jiang}]{vennemeyer2025sycophancythingcausalseparation}
Daniel Vennemeyer, Phan~Anh Duong, Tiffany Zhan, and Tianyu Jiang. 2025.
\newblock \href {https://arxiv.org/abs/2509.21305} {Sycophancy is not one thing: Causal separation of sycophantic behaviors in llms}.
\newblock \emph{Preprint}, arXiv:2509.21305.

\bibitem[{Wartena(2022)}]{wartena-2022-geometry}
Christian Wartena. 2022.
\newblock \href {https://doi.org/10.18653/v1/2022.repl4nlp-1.21} {On the geometry of concreteness}.
\newblock In \emph{Proceedings of the 7th Workshop on Representation Learning for NLP}, pages 204--212, Dublin, Ireland. Association for Computational Linguistics.

\bibitem[{Wartena(2024)}]{wartena-2024-estimating}
Christian Wartena. 2024.
\newblock \href {https://aclanthology.org/2024.konvens-main.9/} {Estimating word concreteness from contextualized embeddings}.
\newblock In \emph{Proceedings of the 20th Conference on Natural Language Processing (KONVENS 2024)}, pages 81--88, Vienna, Austria. Association for Computational Linguistics.

\bibitem[{West and Holcomb(2000)}]{WestHolcomb2000ConcreteAbstractERP}
W.~C. West and P.~J. Holcomb. 2000.
\newblock \href {https://doi.org/10.1162/08989290051137558} {Imaginal, semantic, and surface-level processing of concrete and abstract words: An electrophysiological investigation}.
\newblock \emph{Journal of Cognitive Neuroscience}, 12(6):1024--1037.

\bibitem[{Yang et~al.(2025)Yang, Li, Yang, Zhang, Hui, Zheng, Yu, Gao, Huang, Lv, Zheng, Liu, Zhou, Huang, and others.}]{yang2025qwen3technicalreport}
An~Yang, Anfeng Li, Baosong Yang, Beichen Zhang, Binyuan Hui, Bo~Zheng, Bowen Yu, Chang Gao, Chengen Huang, Chenxu Lv, Chujie Zheng, Dayiheng Liu, Fan Zhou, Fei Huang, and others. 2025.
\newblock \href {https://arxiv.org/abs/2505.09388} {Qwen3 technical report}.
\newblock \emph{Preprint}, arXiv:2505.09388.

\bibitem[{Yao et~al.(2026)Yao, Anand, Zhuang, and Jiang}]{yao2026rhetoricalquestionsllmrepresentations}
Louie~Hong Yao, Vishesh Anand, Yuan Zhuang, and Tianyu Jiang. 2026.
\newblock \href {https://arxiv.org/abs/2604.14128} {Rhetorical questions in llm representations: A linear probing study}.
\newblock \emph{Preprint}, arXiv:2604.14128.

\bibitem[{Zhang et~al.(2023)Zhang, Haddow, and Birch}]{zhang2023promptinglargelanguagemodel}
Biao Zhang, Barry Haddow, and Alexandra Birch. 2023.
\newblock \href {https://arxiv.org/abs/2301.07069} {Prompting large language model for machine translation: A case study}.
\newblock \emph{Preprint}, arXiv:2301.07069.

\bibitem[{Zhuang et~al.(2021)Zhuang, Wayne, Ya, and Jun}]{zhuang-etal-2021-robustly}
Liu Zhuang, Lin Wayne, Shi Ya, and Zhao Jun. 2021.
\newblock \href {https://aclanthology.org/2021.ccl-1.108/} {A robustly optimized {BERT} pre-training approach with post-training}.
\newblock In \emph{Proceedings of the 20th Chinese National Conference on Computational Linguistics}, pages 1218--1227, Huhhot, China. Chinese Information Processing Society of China.

\end{thebibliography}

\appendix

\clearpage

\section{Verb Concreteness Discrimination}
\label{app_verb_conc}

We further extend this analysis to verbs, seeing if the low concrete usage of a verb is identified similarly. We first extract sentences from Wikipedia based on verbs from \citet{concreteness} and train a probe. We then construct a synthetic dataset of action verbs used in a literal vs figurative sense. For instance, ``\textit{He \textbf{grasped} the ball}'' for high-concrete literal usage, ``\textit{He \textbf{grasped} the opportunity}'' for low-concrete figurative usage. 

\begin{figure}[h]
    \centering
    \includegraphics[width=0.99\linewidth]{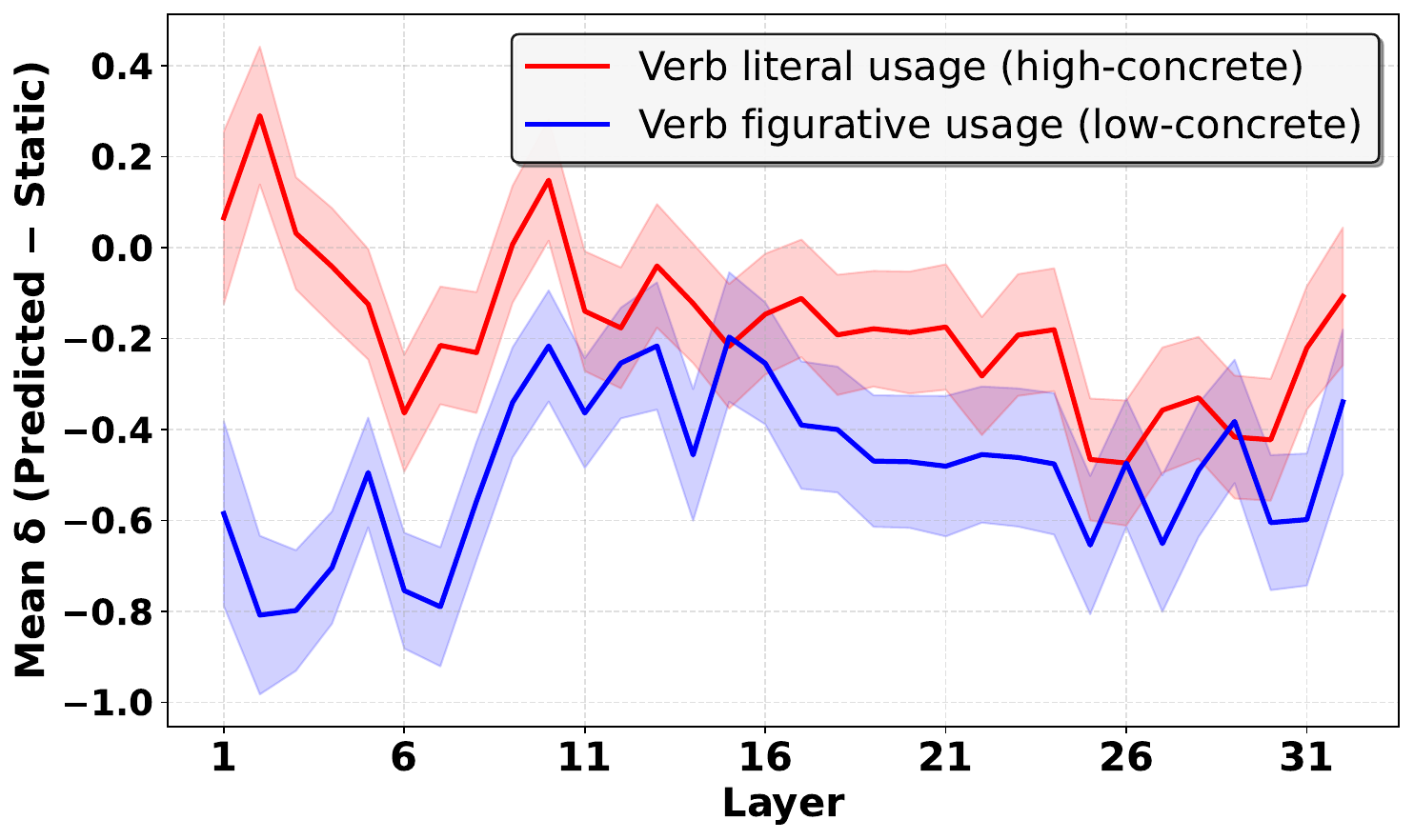}
    \caption{Mean $\delta$ across layers in Llama-3.1-8B, for verbs. Early high separation is followed by moderate to low separation in the middle to later layers.}
    \label{fig:mean_delta_verb}
\end{figure}

Figure~\ref{fig:mean_delta_verb} shows the results for verbs in Llama-3.1-8B. Although the high-concrete verbs have a higher predicted score in all layers, the delta value is not as prominent as nouns. Hence the predicted concreteness scores for literal and figurative verb usages were similar across the layers. These observations agree with previous linguistic work which shows that verbs undergo subtler semantic extensions, while nouns exhibit clearer concreteness shifts between literal and figurative usage~\citep{lakoff1980}. Intuitively, from the example above (\textit{He \textbf{grasped} the ball} vs. \textit{He \textbf{grasped} the opportunity}), we can see that the literal vs. figurative sense is intuitively clear. But the difference in concreteness between the two usages of the word ``\textit{grasped}'' is much less explicit than in noun pairs (\textit{She broke the \textbf{window}} vs \textit{\textbf{window} of opportunity}).

\section{Dataset Distribution}

\begin{figure}[h]
    \centering
    \includegraphics[width=0.99\linewidth]{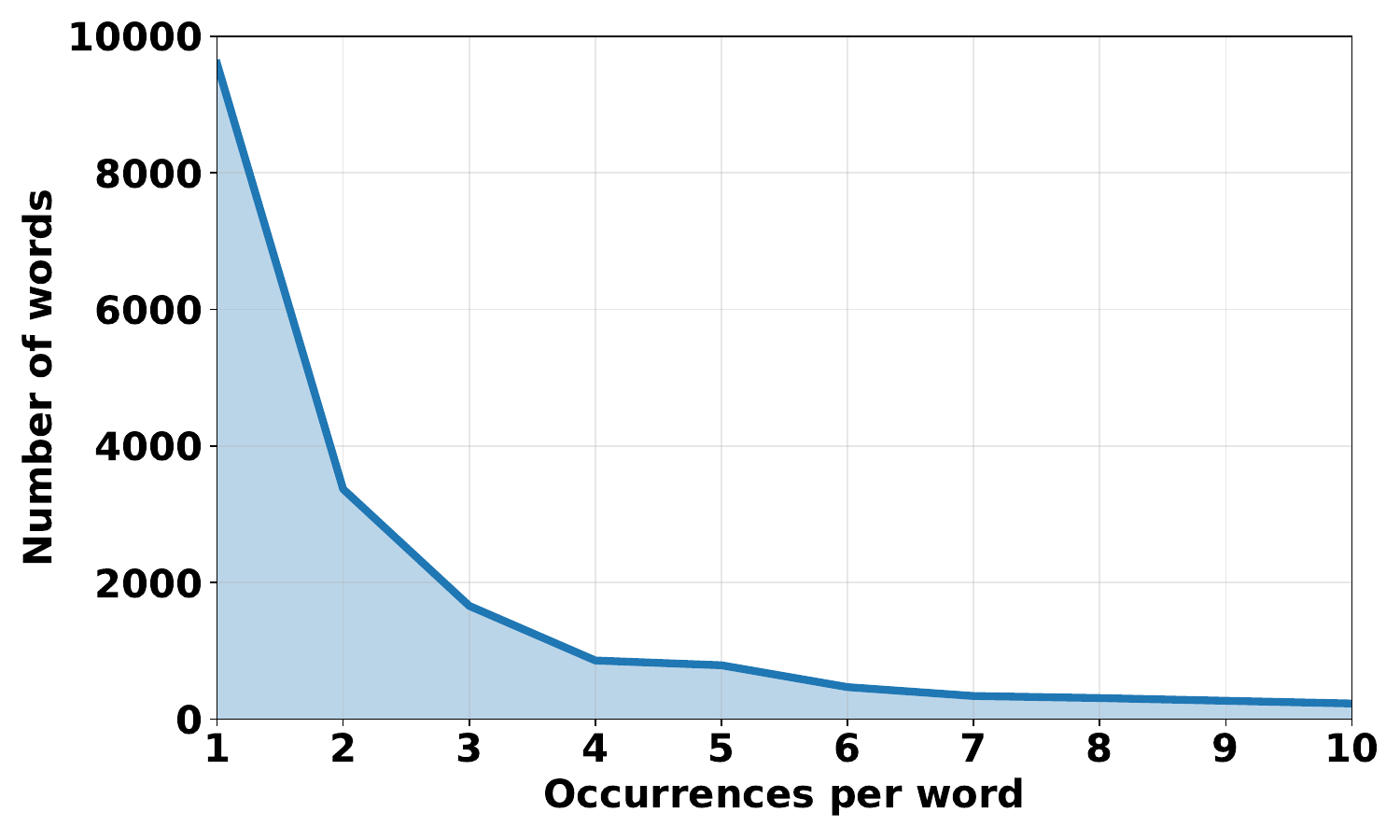}
    \caption{Word frequency distribution in the 25,000 sentences extracted from Wikipedia that we use in our experiments.}
    \label{fig:word_frequency_distribution}
\end{figure}

Figure~\ref{fig:word_frequency_distribution} shows the frequency distribution in the 25,000 sentences extracted from Wikipedia. There are a total of 15,853 unique nouns. The X-axis shows the number of times a word occurs in a sentence in the extracted corpus.

\begin{figure}[h]
    \centering
    \includegraphics[width=0.99\linewidth]{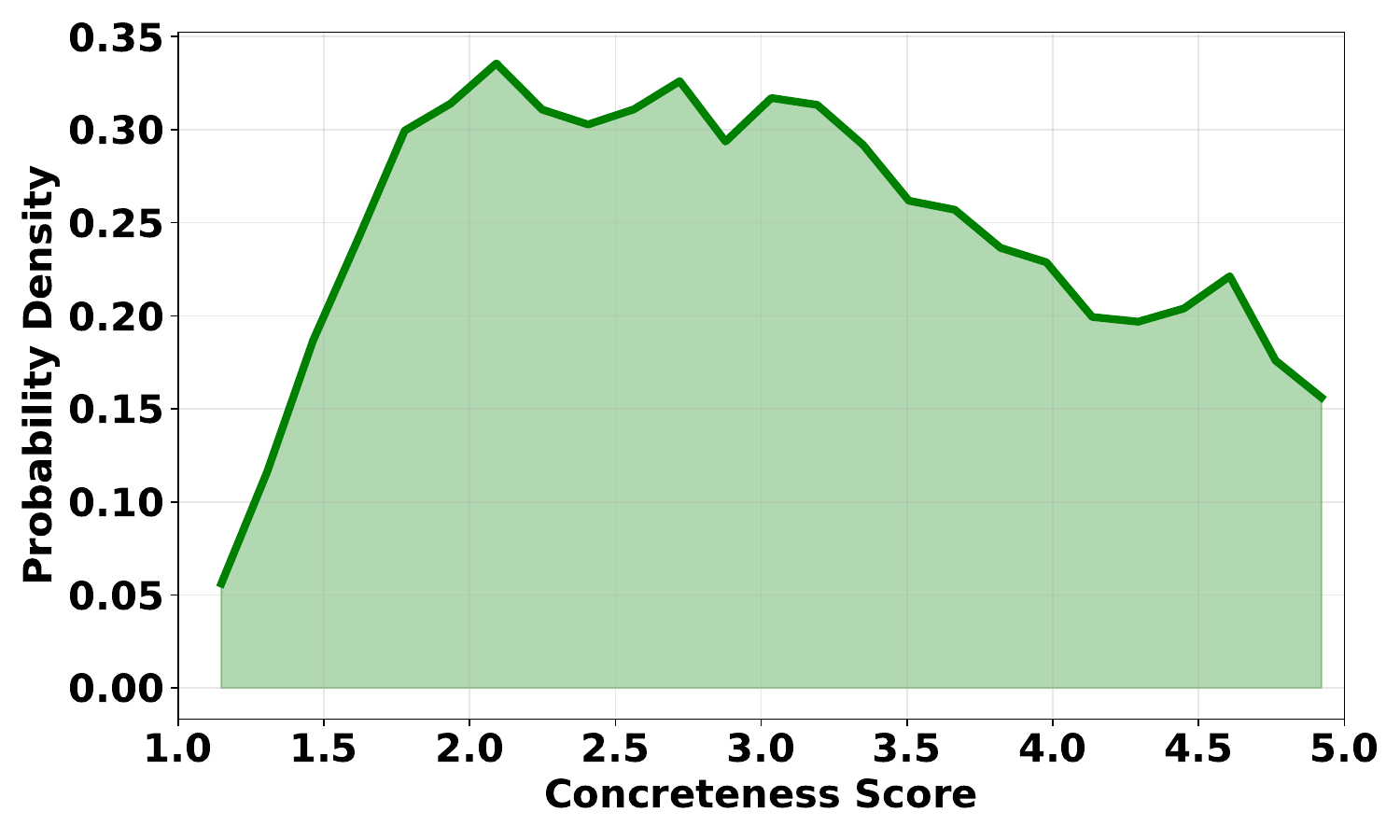}
    \caption{Concreteness score density distribution in the 25,000 sentences extracted from Wikipedia.}
    \label{fig:conc_frequency_distribution}
\end{figure}

Figure~\ref{fig:conc_frequency_distribution} shows the probability density distribution of concreteness scores in the 25,000 extracted sentences from Wikipedia. The distribution shows the concreteness scores being spread out which ensures the diversity of the extracted sentences.

\section{Embedding Correlation Individual Results}
\label{remaining_correlation}

\begin{figure*}[t]
    \centering
    \begin{subfigure}[b]{0.4\linewidth}
        \centering
        \includegraphics[width=\linewidth]{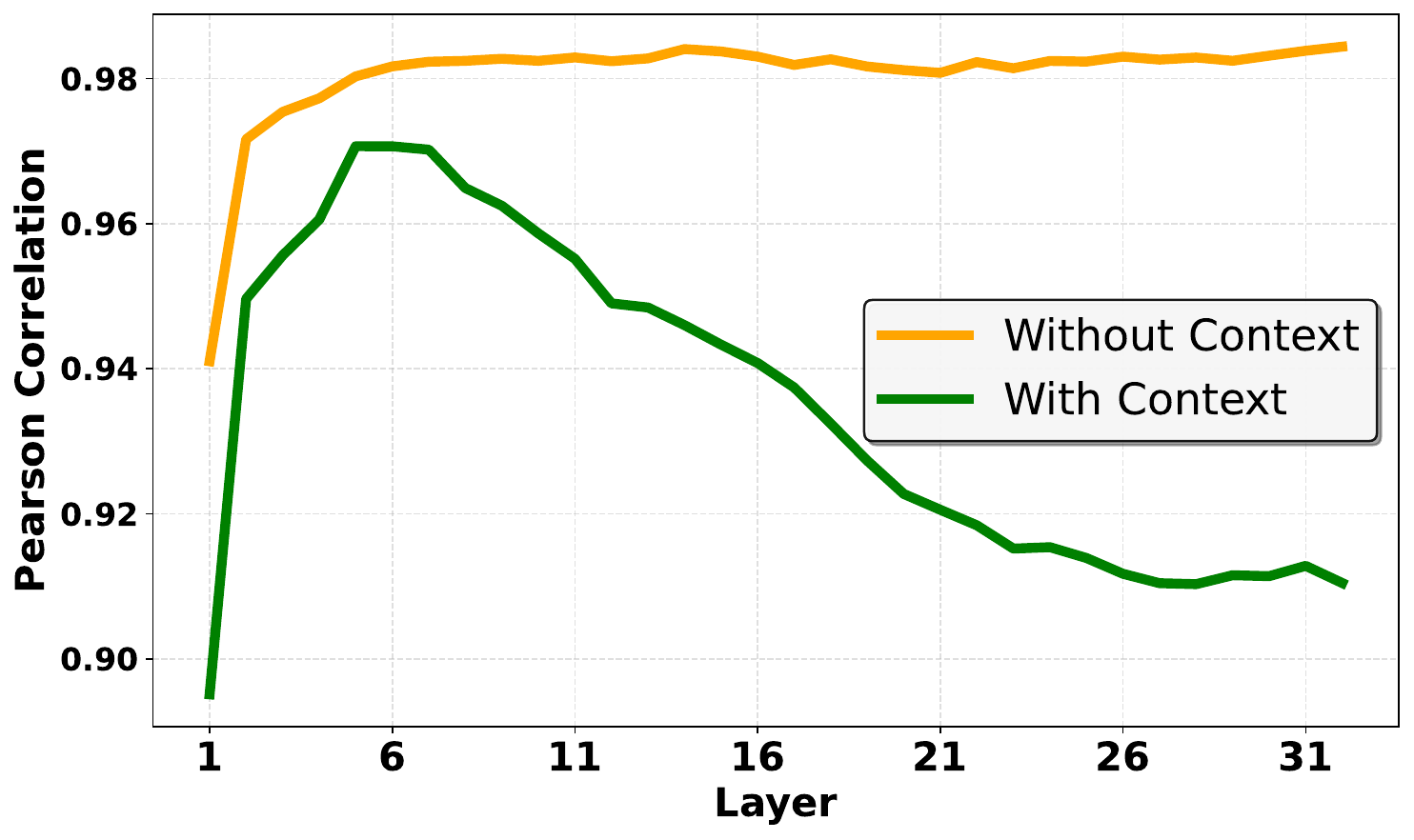}
        \caption{Llama-3.1-8B}
    \end{subfigure}
    \hspace{0.08\linewidth}
    \begin{subfigure}[b]{0.4\linewidth}
        \centering
        \includegraphics[width=\linewidth]{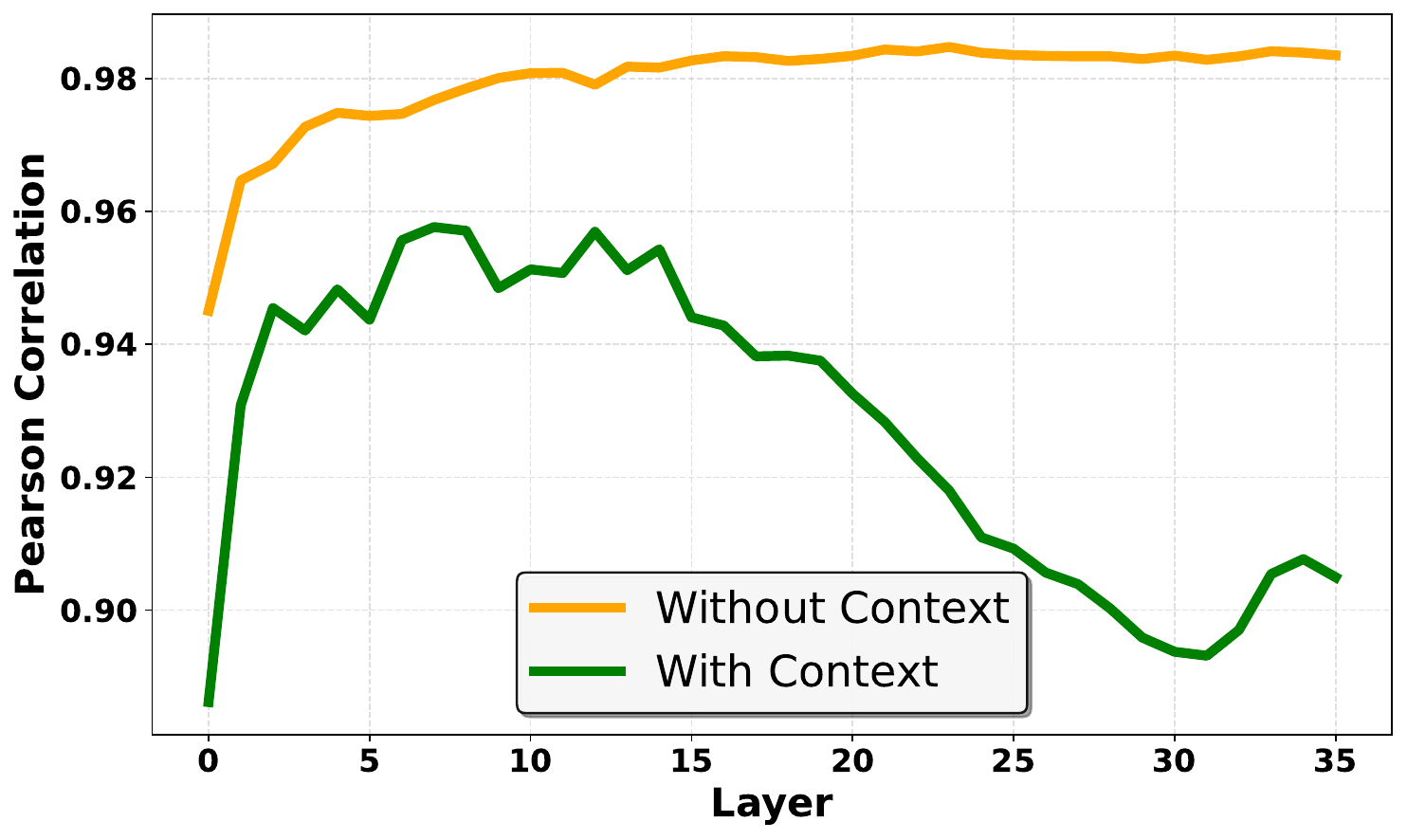}
        \caption{Qwen3-8B}
    \end{subfigure}
    \begin{subfigure}[b]{0.4\linewidth}
        \centering
        \includegraphics[width=\linewidth]{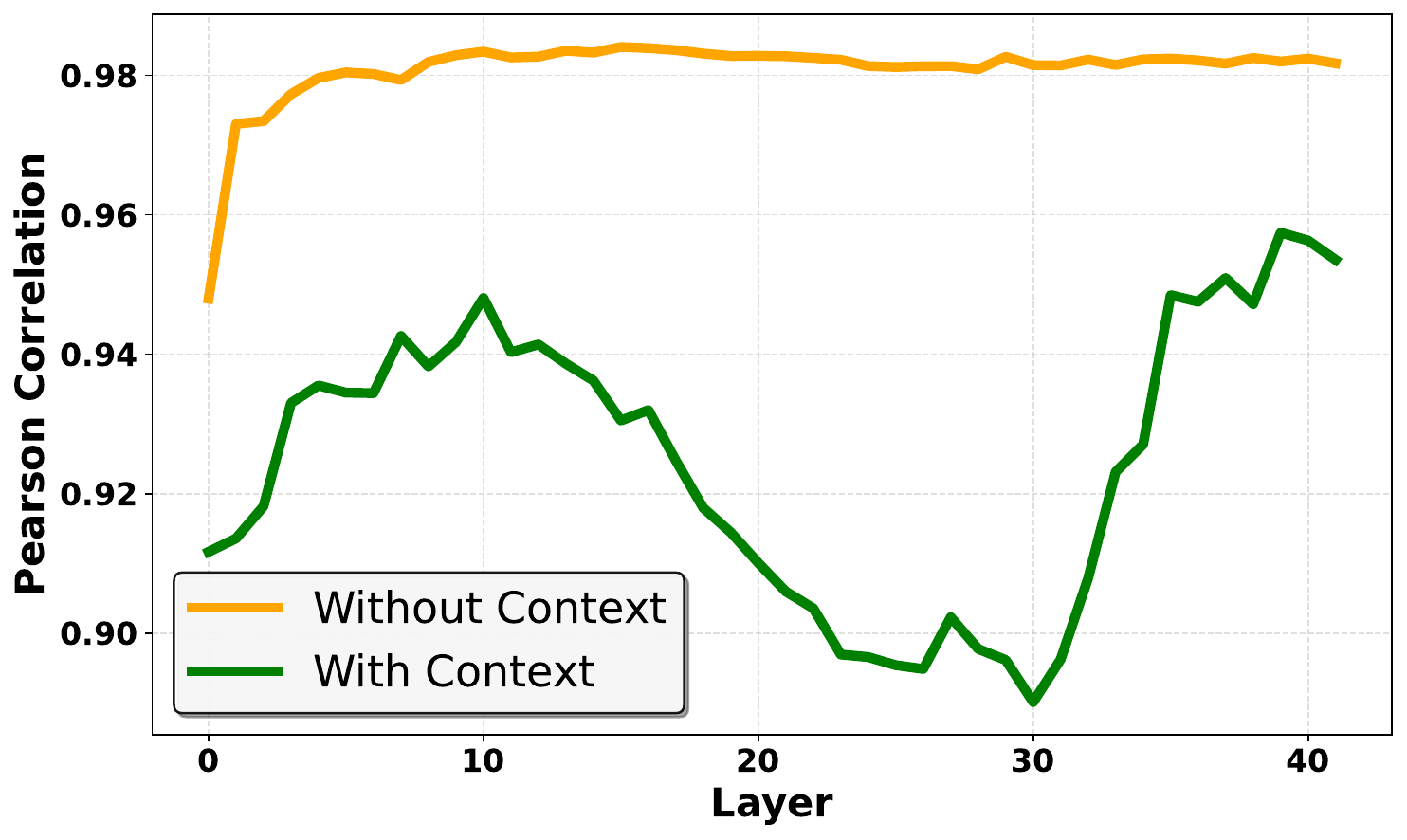}
        \caption{Gemma2-9B}
    \end{subfigure}    
    \hspace{0.08\linewidth}
    \begin{subfigure}[b]{0.4\linewidth}
        \centering
        \includegraphics[width=\linewidth]{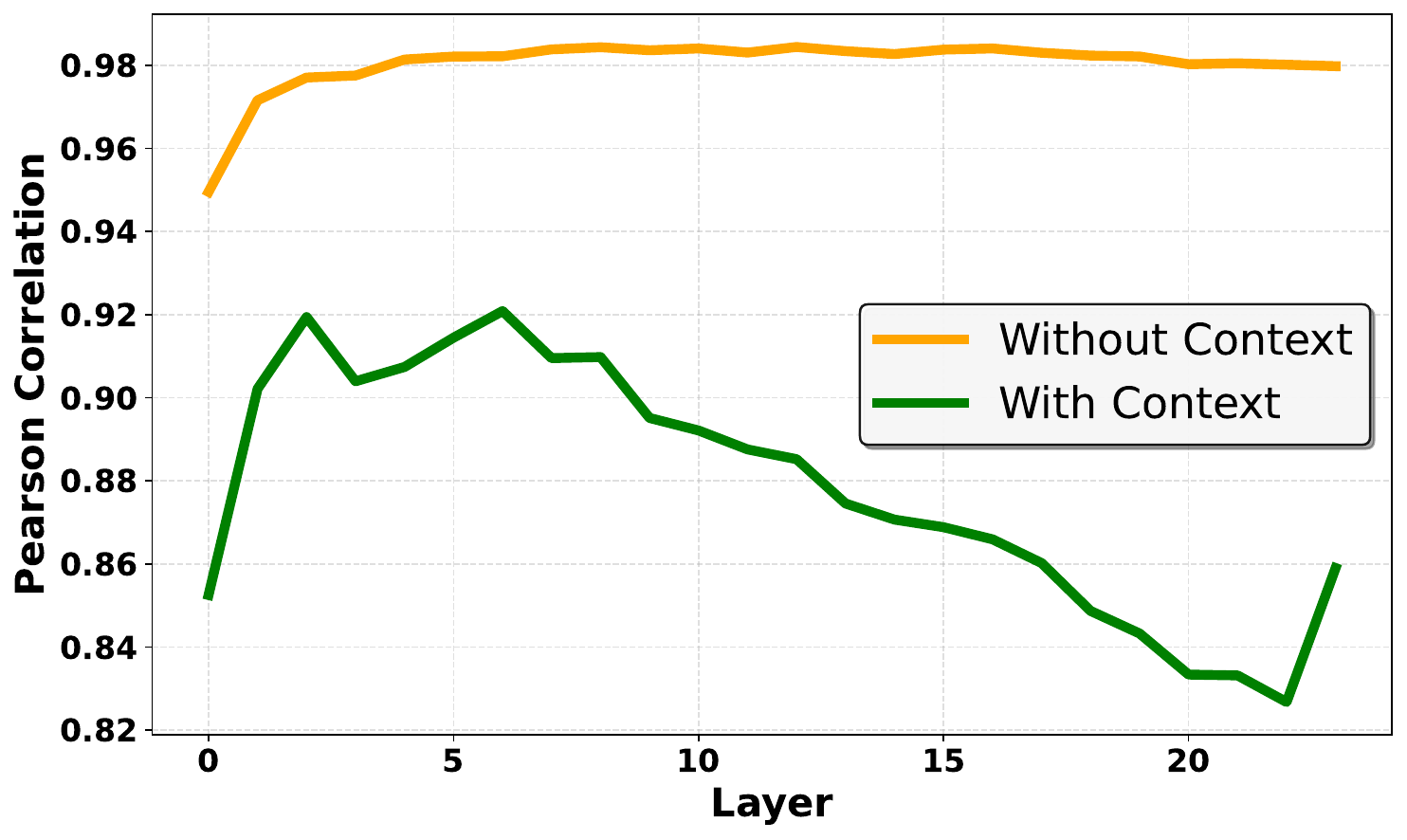}
        \caption{GPT-OSS-20B}
    \end{subfigure}    
    \caption{Pearson correlation between embeddings of models and concreteness scores from \citet{concreteness} for every layer.}
    \label{fig:correlation_remaining}
\end{figure*}

Figure~\ref{fig:correlation_remaining} shows the correlation per layer for static and contextual concreteness for Llama-3.1-8B, Qwen3-8B, Gemma2-9B and GPT-OSS-20B separately. The overall trend remains consistent. Static correlation remains consistently high across all layers. When provided with context, the correlation is highest in the early layers, and then gradually decreases as we move with layers. The trend confirms that all models have similar behavior.

An interesting observation is that the contextual correlation for all models increase in the later layers with varying degree. For Llama-3.1-8B, this rise is the least among the four models. It is more prominent in Qwen3-8B, GPT-OSS-20B and especially in Gemma2-9B. 

\subsection{Correlation Table}
\label{appendix_correlation_table}

\begin{table}[h]
    \centering
    \resizebox{0.8\linewidth}{!}{
    \begin{tabular}{lcc}
    \toprule
    \textbf{Models} & \textbf{Static} & \textbf{Contextual}\\
    \midrule
     Llama-3.1-8B & 0.98 & 0.91 \\
     Qwen3-8B & 0.98 & 0.90  \\
     Gemma-2-9B & 0.98 & 0.92  \\
     GPT-OSS-20B & 0.98 & 0.86  \\
     \bottomrule
    \end{tabular}
    }
\caption{Pearson correlation ($r$) between the last token embedding and concreteness scores using different LLMs using the last hidden layer. The high correlation across all models shows the last token representation carries information about the concreteness of the term.}
\label{table:prompt_based_corr}
\end{table}

Table~\ref{table:prompt_based_corr} reports the Pearson correlations between the predicted scores and human concreteness ratings from \citet{concreteness} for 25,000 sentences from Wikipedia, using only the final layer representation of the last input token. The static setting yields extremely high correlation ($r = 0.98$), confirming that the last token in our prompt-based method robustly captures lexical concreteness. The contextual setting shows slightly lower correlations ($r =0.90$), consistent with the fact that contextual concreteness has a high correlation with static scores, but less than the use of the word in isolation~\citep{Montefinese2023CONcreTEXT}. The results support the validity of our probing method for measuring contextual concreteness in decoder-only LLMs.

\section{Contextual Concreteness Individual Results}
\label{contextual_conc_rem}

\begin{figure*}[t]
    \centering
    \begin{subfigure}[b]{0.32\linewidth}
        \centering
        \includegraphics[width=\linewidth]{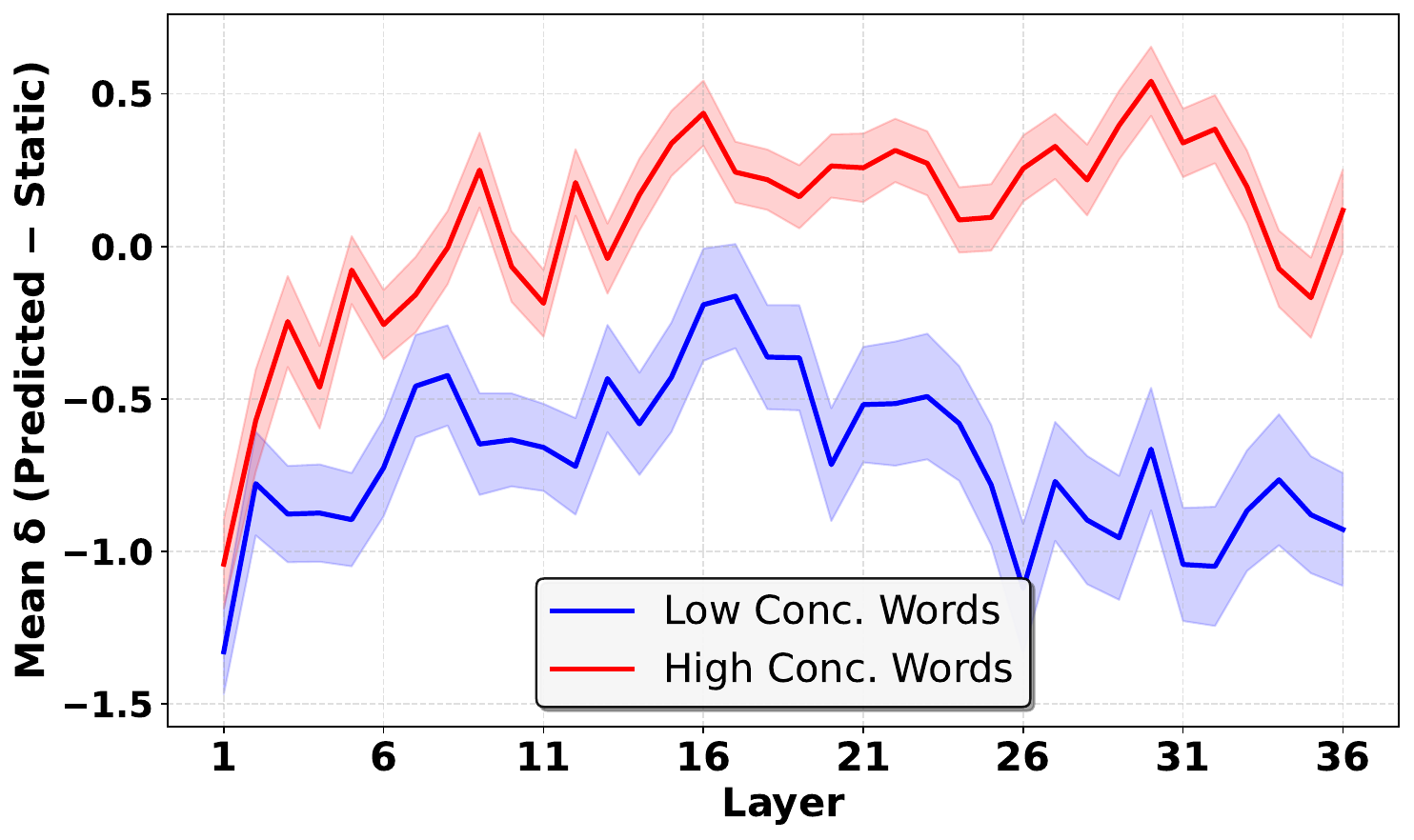}
        \caption{Qwen3-8B}
    \end{subfigure}
    \begin{subfigure}[b]{0.32\linewidth}
        \centering
        \includegraphics[width=\linewidth]{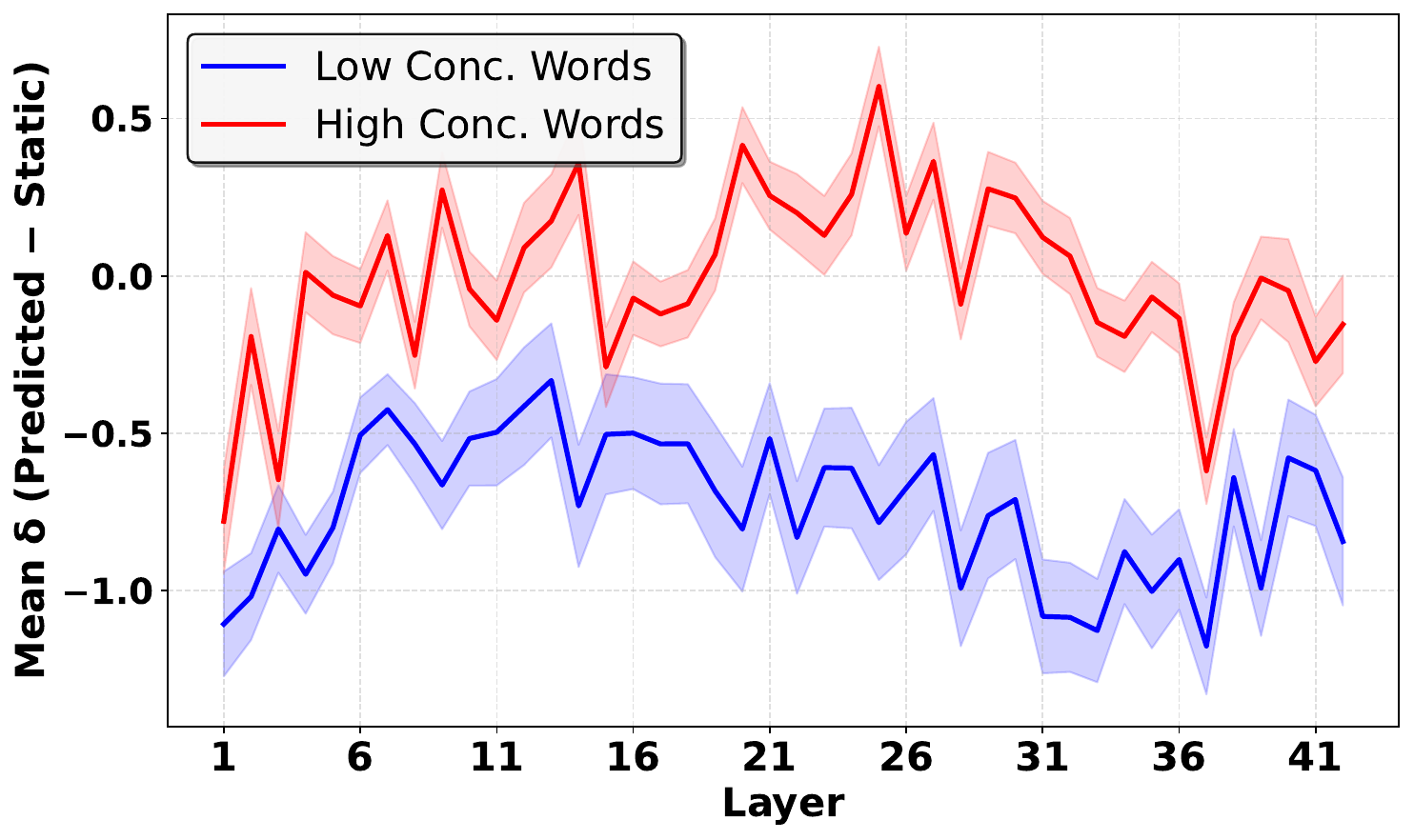}
        \caption{Gemma2-9B}
    \end{subfigure}
        \begin{subfigure}[b]{0.32\linewidth}
        \centering
        \includegraphics[width=\linewidth]{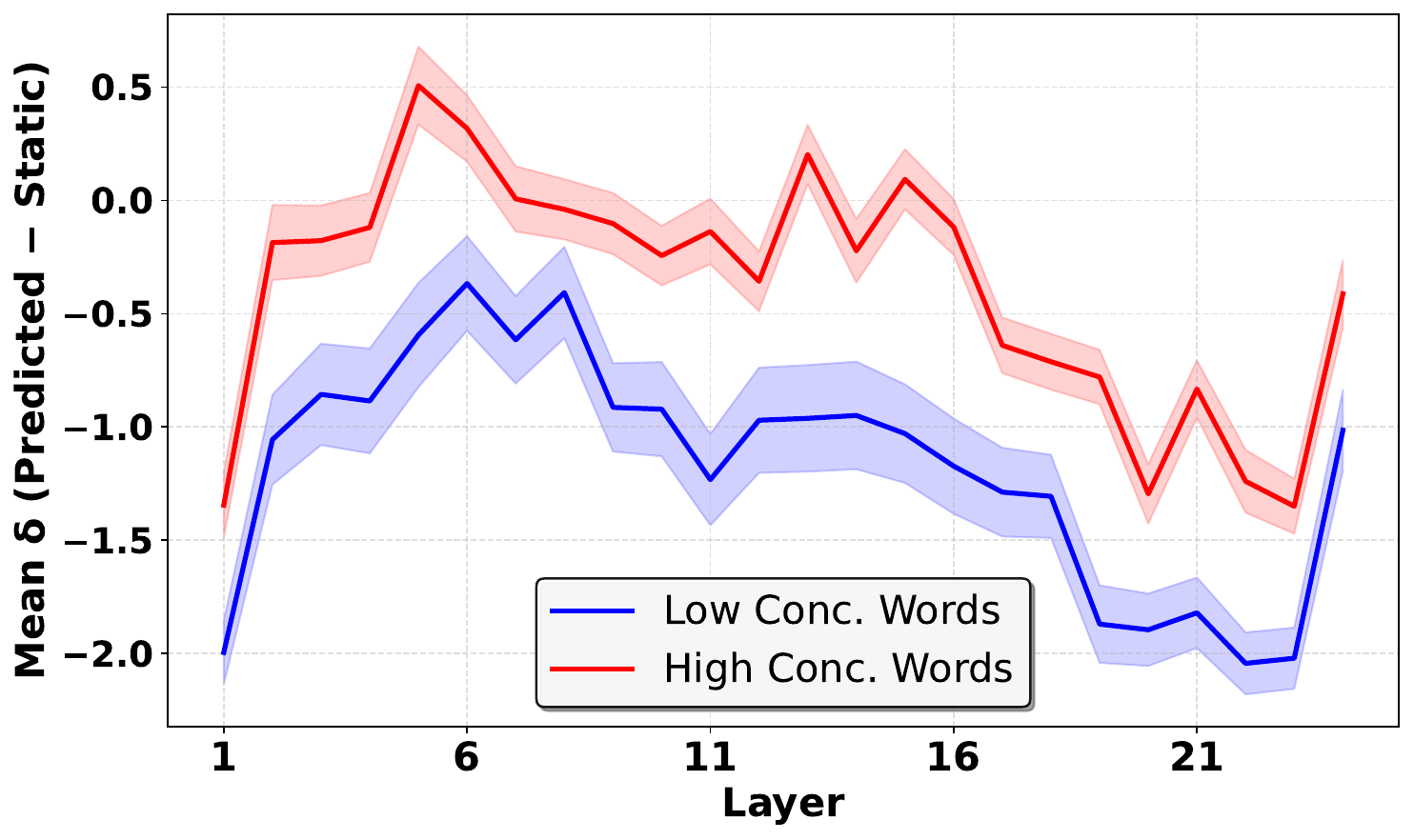}
        \caption{GPT-OSS-20B}
    \end{subfigure}    
    \caption{Mean $\delta$ for all layers for remaining models.}
    \label{fig:mean_delta_2}
\end{figure*}

Figure~\ref{fig:mean_delta_2} shows the layers differentiating high concrete (literal) and low concrete (figurative) usage of the noun for the remaining three models: Qwen3-8B, Gemma2-9B and GPT-OSS-20B. As mentioned in the main body, the behavior of the models are quite similar, with contextual concreteness being discriminated at the very early layers, and remains consistent through all layers.

\section{Geometric Subspaces Individual Results}
\label{appendix_subspace}

\begin{figure*}[t]
    \centering
    \begin{subfigure}[b]{0.4\linewidth}
        \centering
        \includegraphics[width=\linewidth]{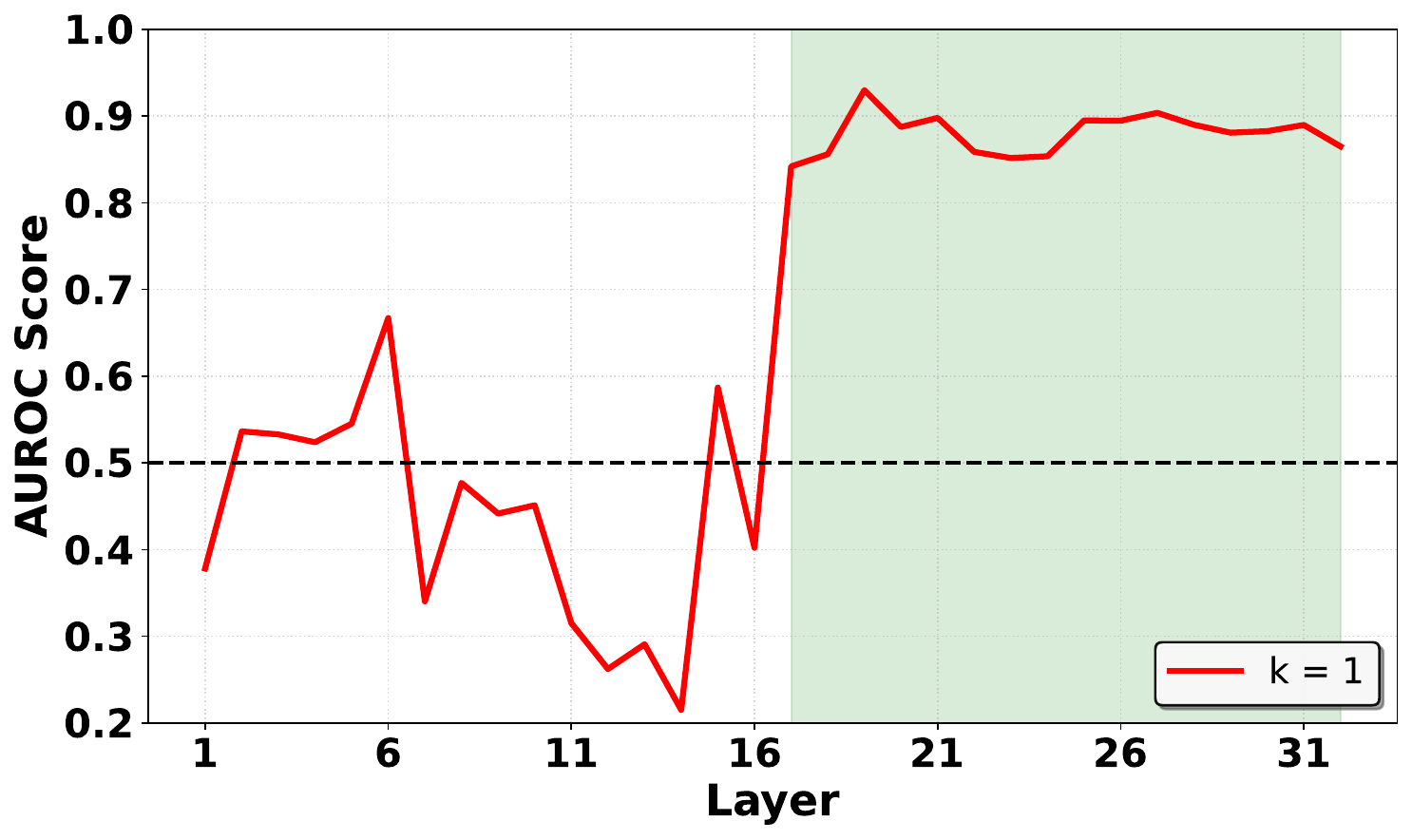}
        \caption{Llama-3.1-8B}
    \end{subfigure}
    \hspace{0.08\linewidth}
    \begin{subfigure}[b]{0.4\linewidth}
        \centering
        \includegraphics[width=\linewidth]{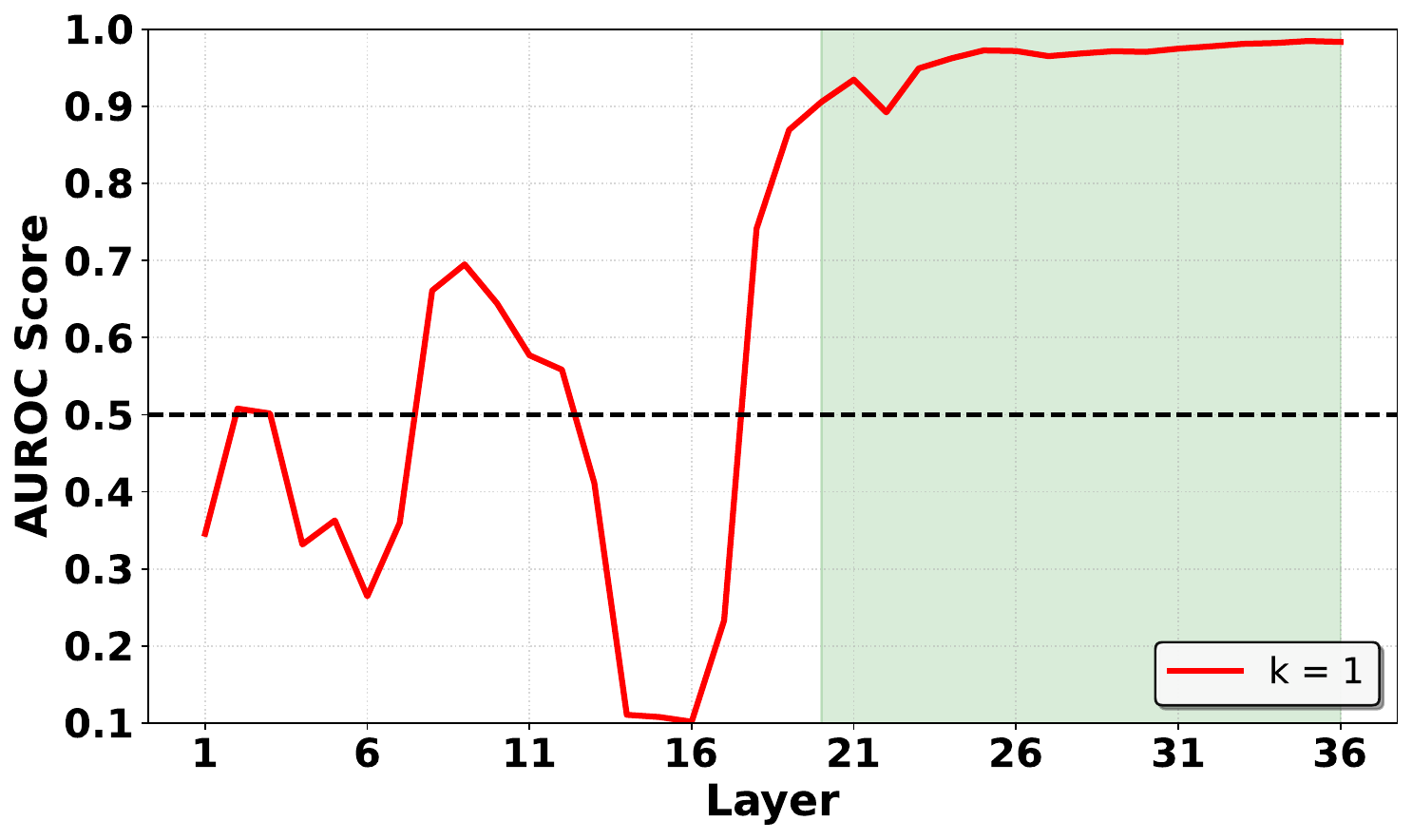}
        \caption{Qwen3-8B}
    \end{subfigure}    
    \begin{subfigure}[b]{0.4\linewidth}
        \centering
        \includegraphics[width=\linewidth]{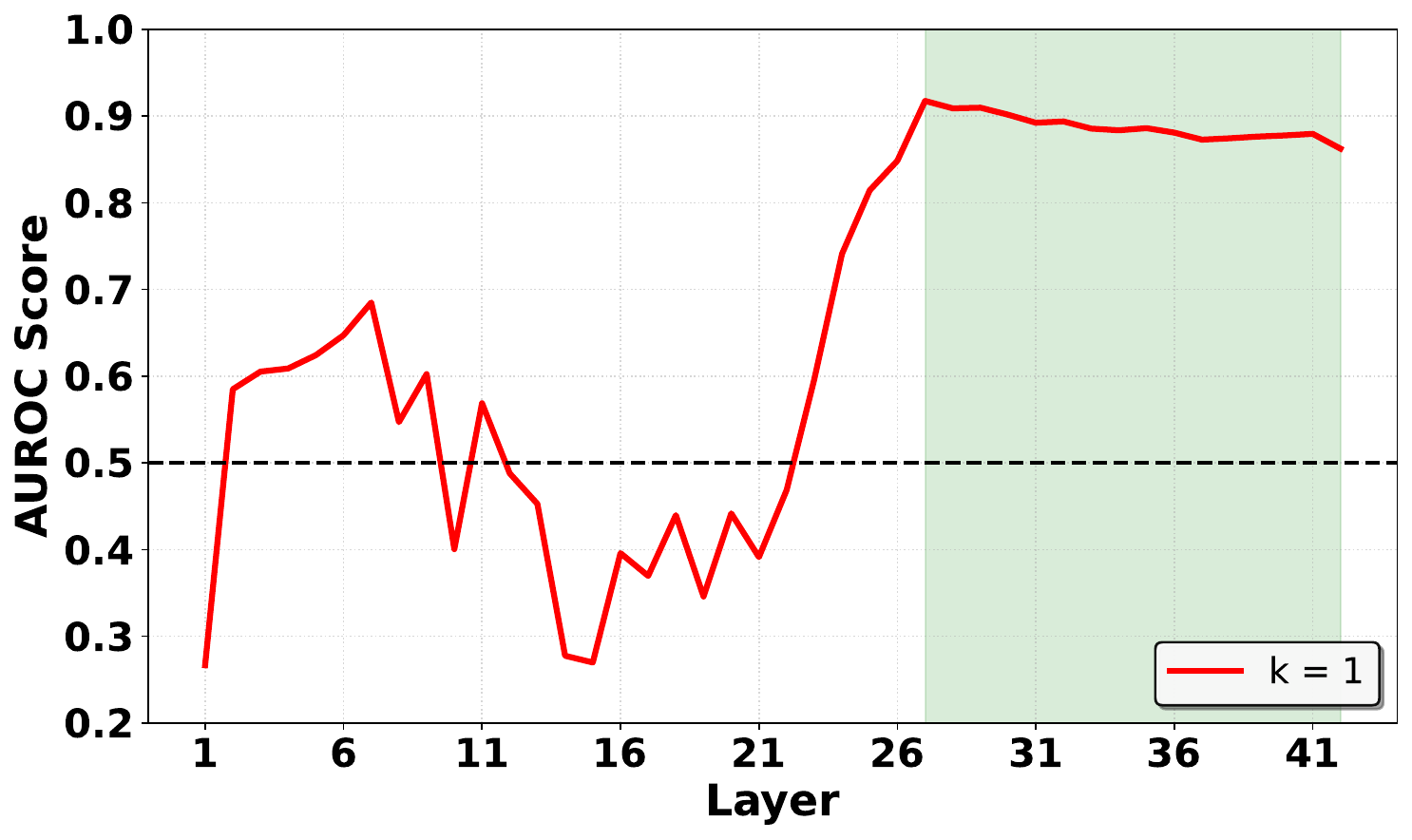}
        \caption{Gemma2-9B}
    \end{subfigure}
    \hspace{0.08\linewidth}
        \begin{subfigure}[b]{0.4\linewidth}
        \centering
        \includegraphics[width=\linewidth]{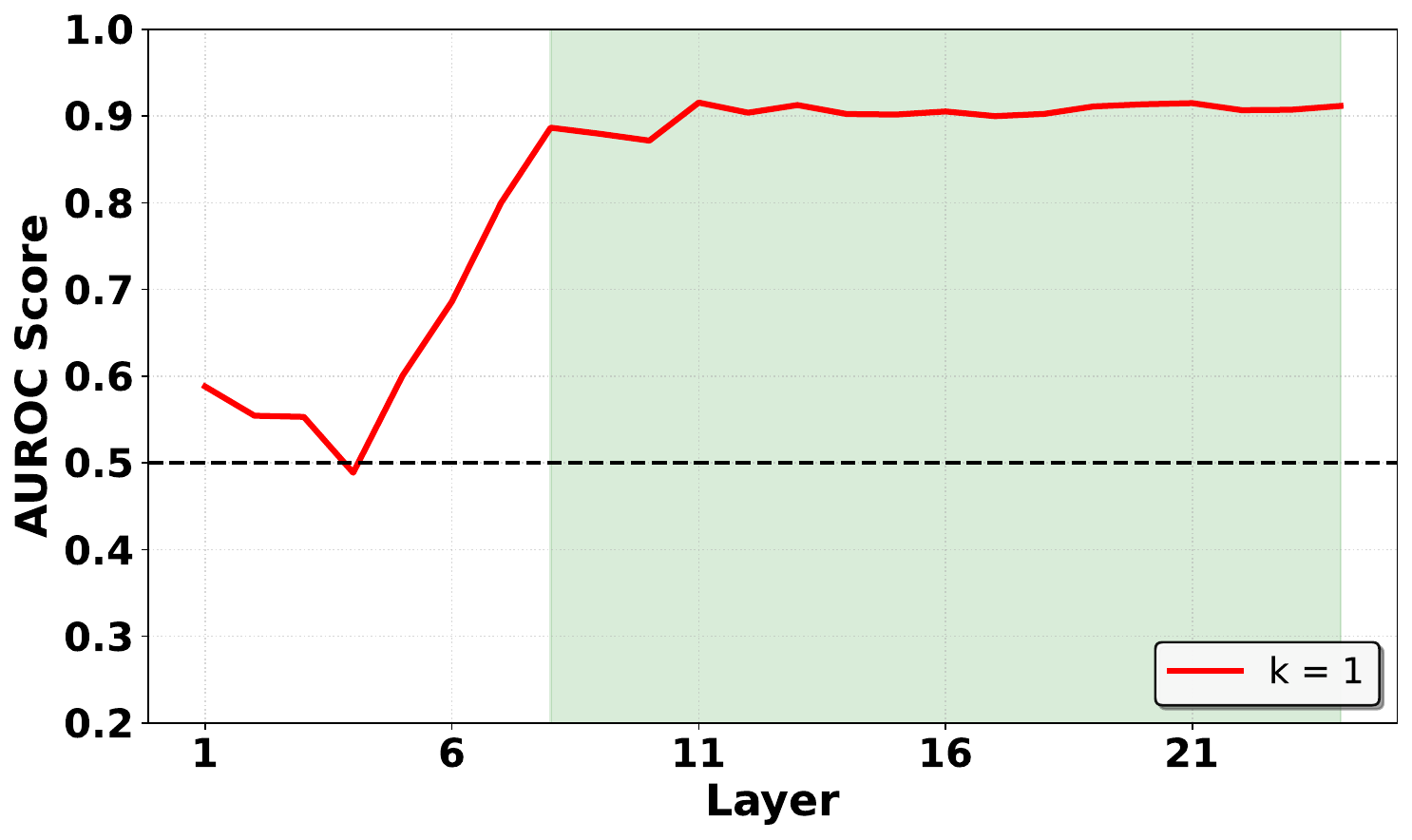}
        \caption{GPT-OSS-20B}
    \end{subfigure}    
    \caption{AUROC score for classifying high and low concrete nouns using one-directional geometric subspace for remaining models.}
    \label{fig:subspace_2}
\end{figure*}

Figure~\ref{fig:subspace_2} shows the one-directional subspace performance in differentiating high-concrete (literal) and low-concrete (figurative) usage of the noun for the remaining models: Qwen3-8B, Gemma2-9B and GPT-OSS-20B. Like the results in the main body, we see that concreteness is compressed into a linear subspace universally in the later layers, with AUROC scores around 0.9. The one-directional compression for Qwen3-8B and Gemma2-9B is similar to Llama, starting around the late-mid layers. For GPT-OSS-20B the compression begins in the early-mid layers. We believe GPT-OSS shows slightly different behavior due to its mixture-of-expert architecture~\citep{lo2025closerlookmixtureofexpertslarge}. The results confirm that LLMs universally compress concreteness into a singular directional subspace in the later layers.

\subsection{Effect of Subspace Dimensionality}

\begin{figure}[h]
    \centering
    \includegraphics[width=0.98\linewidth]{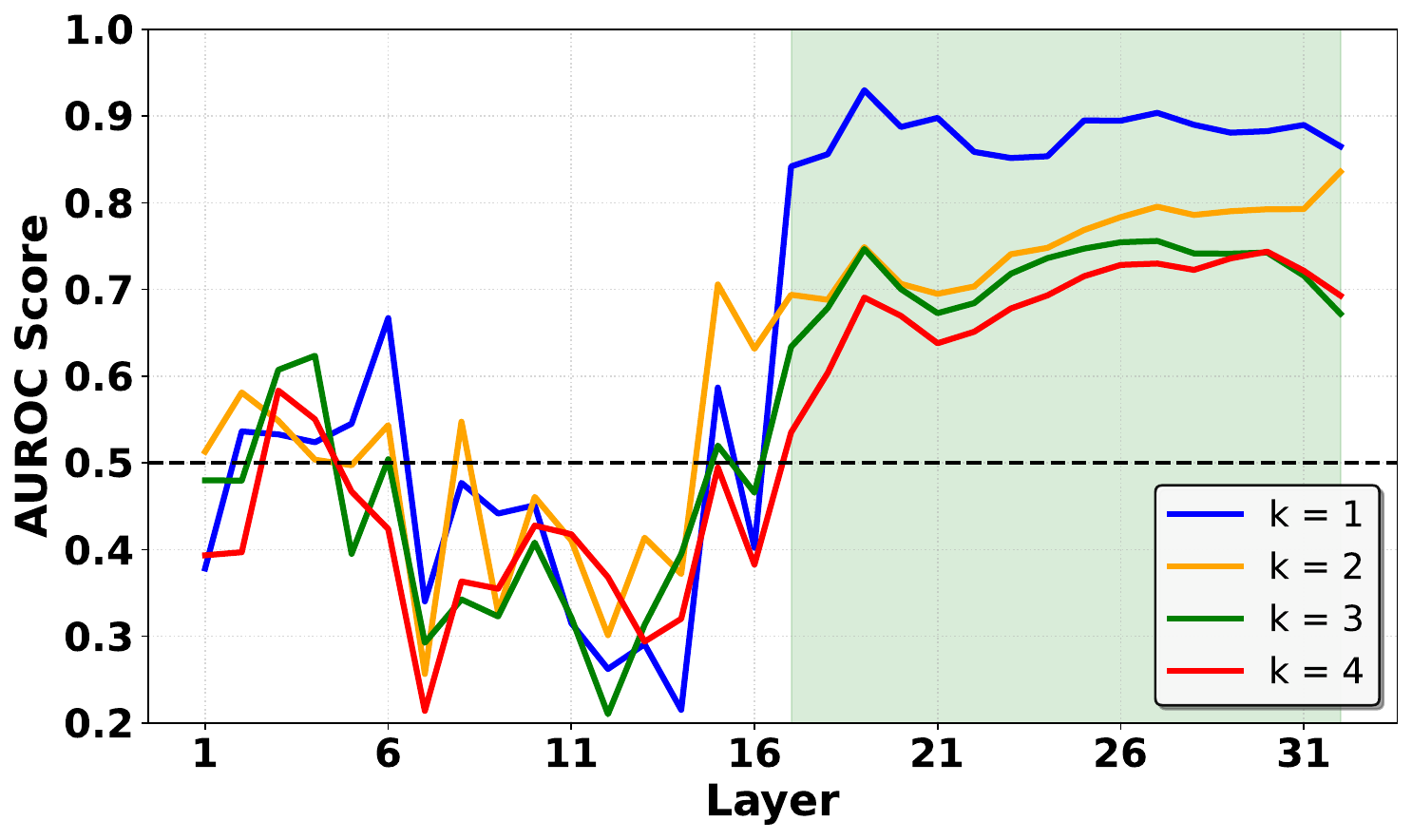}
    \caption{AUROC score for classifying high and low concrete nouns using $k=1,2,3,4$ for Llama-3.1-8B.}
    \label{fig:subspace_123}
\end{figure}

We further examine the effect of increasing the dimensionality of the concreteness subspace. Figure~\ref{fig:subspace_123} reports results for $k$ = 2, 3, and 4 using Llama-3.1-8B; we observe the same qualitative trends across all evaluated models. As the subspace dimensionality increases, the AUROC scores in the middle through to the later layers consistently decrease. This degradation indicates that the concreteness signal becomes less discriminative when spread across multiple dimensions, rather than being aligned along a single dominant direction.

This behavior provides additional evidence that concreteness is primarily encoded along a unidirectional axis in deeper layers. While higher-dimensional subspaces can capture more total variance, they also incorporate secondary directions that are less consistently associated with concreteness and may reflect noise or task-irrelevant variation. Consequently, expanding the subspace dilutes the discriminative signal, leading to weaker separation between literal and figurative usages. These findings further support our central claim that concreteness is compressed into a single, coherent direction in the later layers of large language models, and that this compact representation is most effective for downstream probing and control.

\section{Figurative Text Datasets}
\label{figurative_text_dataset}

In this section, we describe how we process the datasets for our experiments. Our objective is to extract sentences where the figurative meaning is realized by a noun.

\noindent \textbf{Metaphor.} We use two human-annotated metaphor datasets: VUAMC (VUA)~\citep{vua} and MUNCH~\citep{munch}. We chose these two datasets because both datasets provide token-level metaphor annotations, indicating which word or phrase in a sentence is used metaphorically. We first identify the metaphor-annotated token and then apply part-of-speech tagging using SpaCy to determine whether the metaphorically used word is a noun. If so, we extract the sentence as a metaphorical sample and record the noun as the target word. From each of VUA and MUNCH, we collect 1,000 such noun-metaphor sentences. To construct a balanced dataset for classification, we additionally sample 1,000 noun usages where the target noun is annotated as \textit{non-metaphorical}. This results in an equal distribution of metaphor and literal examples.

\noindent \textbf{Idioms.} We use two human-annotated idiom datasets: EPIE~\citep{saxena2020epiedatasetcorpuspossible} and MAGPIE~\citep{haagsma-etal-2020-magpie}. Just like the metaphor datasets, both these datasets include span-level annotations that mark idiomatic expressions within sentences. Following the same pipeline as in the metaphor setting, we identify the annotated idiomatic span and apply SpaCy POS analysis to determine whether the idiomatic expression involves a noun. Sentences meeting this criterion are extracted as positive idiom samples, with the noun recorded as the target word. We collect 1,000 idiomatic sentences from each dataset. To form a balanced dataset, we additionally sample 1,000 noun usages from instances that are explicitly marked as non-idiomatic, yielding an equal distribution of positive versus negative examples.

\noindent \textbf{Metonymy.} For metonymy, we use the ConMeC~\citep{ghosh-jiang-2025-conmec} and MetFuse~\citep{ghosh2026metfusefigurativefusionmetonymy}, both of which are human-annotated resources. ConMeC has instances of common noun metonymy spread across six categories. MetFuse contains a combination of common noun and named entity metonymy, spread across locational or institution based metonymic mapping. In these datasets, the metonymic shift is realized through a noun, and the target noun is explicitly provided alongside the sentence. We sample 1,000 metonymic and 1,000 literal noun instances from ConMeC to obtain a balanced set. MetFuse consists of 1,000 parallel sentences consisting of literal, metonymic, metaphoric and hybrid expressions. We incorporate all the samples, using the metonymic sentences as positive and literal sentences as negative cases.

\section{Token Generation Prompt \& MLP Details}
\label{token_generation_prompt}

\subsection{Prompts Used}

\begin{figure}[h]
\centering
\noindent
\fbox{%
    \parbox{0.9\linewidth}{%

        \medskip
        sentence: \textit{[sentence]}

        \medskip
        On a scale of 1 to 5 (5 being the highest), in the context of the sentence, 
        what is the concreteness of the word ``\textit{[target\_word]}''?
    }%
}
\caption{Prompt for generating contextual concreteness token. The last token representation is used to carries information regarding the concreteness of the term.}
\label{fig:contextual_prompt}
\end{figure}

\begin{figure}[h]
\centering
\noindent
\fbox{%
    \parbox{0.97\linewidth}{%
        \textbf{Static Concreteness Token Generation Prompt:}

        \medskip
        On a scale of 1 to 5 (5 being the highest), what is the concreteness of the word \textit{[target\_word]}?
    }%
}
\caption{Prompt for generating static concreteness token.}
\label{fig:static_prompt}
\end{figure}

Figure~\ref{fig:static_prompt} shows the prompt used for static concreteness information. We simply provide the target word without context. Figure~\ref{fig:contextual_prompt} shows the prompt used to generate the a singular token that carries the contextual concreteness information in our prompt-based contextual probing technique. We provide the sentence, and then ask the model to predict the concreteness score of the target word in context of that sentence. 

\subsection{Prompt Sensitivity Analysis}

We perform further prompt sensitivity analysis as our experiments show that the structure of the prompt affects the concreteness information in the last token representation. 

While our prompt-based probing method reliably captures contextual concreteness, we observe sensitivity to the placement of the target noun within the prompt structure. When the target noun appears as the \textit{final} lexical item in the prompt, the predicted concreteness scores show extremely high correlation with human norms (Pearson $r \approx 0.98$). In contrast, when the noun occurs earlier in the sentence—followed by several contextual tokens—the correlation decreases ($r \approx 0.80\pm0.10$). This suggests that decoder models prioritize the most recent context in their internal representations during generation, consistent with recency-weighted attention and local next-token prediction behavior. Similar effects have been reported in studies showing that token salience increases toward the rightmost position during decoding in autoregressive LLMs, with downstream tasks benefiting when key information is placed near the model’s prediction point \citep{mialon2023augmentedlanguagemodelssurvey, zhang2023promptinglargelanguagemodel, brown2020languagemodelsfewshotlearners}. Our prompt was designed accordingly and we selected the prompt that had the maximum correlation score for our experiments.

\subsection{MLP Details}

Our MLP follows the same settings that of \citet{wartena-2024-estimating}. It consists of three hidden layers (dimensions: 512~$\rightarrow$~256~$\rightarrow$~128), with ReLU activations and a dropout rate of 20\% at each hidden layer. AdamW optimizer was used with a learning rate of $1\times 10^{-5}$, weight decay of $1\times 10^{-4}$, trained for 50~epochs using a batch size of 15. Using this trained regression model, we predict the concreteness score of the target noun using 10-fold cross-validation. In each fold, the model is trained on 9 folds and used to predict concreteness scores on the held-out fold. We compute Pearson correlations between the predicted and human-rated concreteness values from \citet{concreteness}, averaged across folds.

\section{Synthetic Data Generation}
\label{synthetic_data_generation}

\subsection{Prompt Used}

\begin{figure*}[t]
\centering
\noindent
\fbox{%
    \parbox{0.97\linewidth}{%
        \textbf{Synthetic Data Generation Prompt}

        \medskip
        \textit{System Prompt:} You are a helpful assistant. You are given an English word that is concrete but can also be used abstractly depending on context. Here is your task:\\
        1. Generate a sentence where this word is used in a highly concrete (or sensory) way.\\
        2. Generate a sentence where this word is used in a less concrete (figurative) way.

        \medskip
        \textit{Few-shot exemplar 1}\\
        Word: city\\
        highly concrete = The \textit{city} is 5 km away from this location.\\
        less concrete = He was overwhelmed by the \textit{city}.

        \medskip
        \textit{Few-shot exemplar 2}\\
        Word: window\\
        highly concrete = she opened the \textit{window} as it was getting hot.\\
        less concrete = she saw a \textit{window} of opportunity and pounced

        \medskip
        \textit{Few-shot exemplar 3}\\
        Word: door\\
        highly concrete = He painted the wooden \textit{door}.\\
        less concrete = He closed the \textit{door} on their relationship.
        
        \medskip
        \textit{Input}\\
        Word: \textit{[target\_word]}
    }%
}
\caption{Prompt for generating static concreteness token.}
\label{fig:synthetic_prompt}
\end{figure*}

Figure~\ref{fig:synthetic_prompt} shows the prompt used to generate the synthetic data using GPT-5.1. We use three few-shot exemplars of metonymic, metaphorical and idiomatic example.

\subsection{Annotation Details}

For the annotation task, we recruited students and colleagues. Their participation was completely voluntary. Figure~\ref{fig:annotation_guidelines} provides the annotation guidelines provided to the annotators. For the synthetic dataset annotation, two annotators judged 600 high-concrete and 600 low-concrete sentences. All 600 sentences were judged to be literal by both the annotators, and 572 were judged to have the noun used in a low-concrete sense. We identified that 14 sentences were marked to low-concrete in a word-sense disambiguation sense and not in a figurative sense. We removed those cases from the final pool of 500 sentences to ensure the dataset is consistent.

For the steering task, we used two different human annotators to avoid bias. The guidelines were different, the annotators were asked to judge if the sentence as a whole is literal or figurative and not focus on an individual noun. 

\begin{figure*}[t]
\centering
\noindent
\fbox{%
    \parbox{0.97\linewidth}{%
        \textbf{Annotation guidelines}

        \medskip
        Concreteness denotes how physical or abstract a term is. Physical entities that can be perceived or touched are \textit{concrete} terms, while words referring to intangible or abstract notions are less-concrete (or abstract term).\\
        Your task is binary classification. You will be given a sample, each sample contains a pair of sentences where the \textit{noun} is used in a different way, highly-concrete or less-concrete than its usual norm. The noun is usually highly-concrete in isolation, but it can also be used in a low-concrete sense when used figuratively. The pair of sentences in each sample are:

        \medskip
        \textbf{1. Literal Sentence.} The noun is used in its usual high-concrete sense.\\
        Example: \textit{He climbed the \textbf{ladder} to reach the roof}. Here, \textit{ladder} refers to a physical ladder, and is therefore used in its usual high-concrete sense.

        \medskip
        \textbf{2. Figurative Sentence.} The noun is used in a high-concrete sense than its lexical norm, mainly in a figurative sense.\\
        Example: \textit{He climbed the corporate \textbf{ladder} to the position of VP}. Here, \textit{ladder} does not refer to its physical self, but used in a metaphorical sense, therefore is less-concrete than its lexical norm.\\
        (Note that the term does not have to be abstract. It just needs to be used in a sense than its usual norm).
        
        \medskip
        \textbf{Primary Guidelines:}\\
        1. If the noun carries its original concrete sense in the literal sentence, it is considered correct (mark as 1, else 0).\\
        2. If the noun is used in a \textit{less-concrete} sense in the supposedly figurative sentence, it is considered correct (mark as 1, else 0).\\

        \medskip
        \textbf{Additional Guidelines:}\\
        The intended figurative expressions are primarily metaphors, metonymy and idiom. If the noun is used in a less-concrete sense through word-sense disambiguation (eg. - square \textit{root} of 7), please mark that sentence.\\
        
    }%
}
\caption{Annotation guidelines.}
\label{fig:annotation_guidelines}
\end{figure*}

\section{Additional Steering Examples \& Discussion}
\label{app_steering_example}

Table~\ref{tab:fig-to-lit} are examples of steering from figurative to literal. Interestingly, we observed that when using the steering method, the literal examples \textit{resolved} the figurative instance rather than write it write it out, which was more common in the general prompt without steering. For example, in ``\textit{The \textbf{city} attacked the new policy,}'' the metonymy occurs through the noun \textit{city}. The steered output directly resolves the figurative usage of the noun, rendering it literal. Therefore, the rewrites of the steered generation tended to resolve the figurative noun, while the non-steered generation tended to rewrite the sentence to get rid of the figurative instance.

Table~\ref{tab:lit-to-fig} are examples of steering from literal to figurative. As we mentioned earlier, previous work have shown that LLMs find it hard to go in this direction. Higher $\alpha$ values often led to unwanted noise of randomness.

\begingroup
\renewcommand\baselinestretch{0.99}
\begin{table*}[t]
    \NiceMatrixOptions {
    custom-line = {
       command = dashedmidrule ,
       tikz = { dashed } ,
       total-width = \pgflinewidth + \aboverulesep + \belowrulesep ,
     } }
    \centering
    \small
    \resizebox{\textwidth}{!}{
    \begin{NiceTabular}{|p{0.33\linewidth}|p{0.33\linewidth}|p{0.33\linewidth}|}
    \toprule
    \multicolumn{1}{c}{\bf Original Figurative Sentence} & \multicolumn{1}{c}{\bf Unsteered Rewrite} & \multicolumn{1}{c}{\bf Steered Literal Rewrite} \\
    \midrule
    Her apology served as a bridge to mend their broken relationship. & Her apology helped mend their broken relationship. \tg{\textsf{\textbf{F}}} & Her apology provided a way to heal their relationship. \tg{\textsf{\textbf{F}}} \\ 
    \midrule
    His actions served as a mirror to his true intentions. & His actions reflected his true intentions. \tg{\textsf{\textbf{F}}} & His actions revealed his true intentions. \tg{\textsf{\textbf{F}}} \\ 
    \midrule
    She was the gatekeeper of her family's secrets. & She stood as the gatekeeper of her family’s secrets. \tr{\textsf{\textbf{L}}} & Her family's secrets were her responsibility to keep. \tg{\textsf{\textbf{F}}} \\ 
    \midrule
    The city attacked the new policy. & The city strongly opposed the new policy. \tr{\textsf{\textbf{L}}} & The city's residents and businesses were against the new policy. \tg{\textsf{\textbf{F}}}\\ 
    \bottomrule
    
    \end{NiceTabular}
    }
    \caption{Example of concreteness guided steering for Figurative $\rightarrow$ Literal. \tr{\textsf{\textbf{L}}} indicates the rewritten sentence was literal and \tg{\textsf{\textbf{F}}} means figurative.}
    \label{tab:fig-to-lit}
\end{table*}
\endgroup

\section{Figurative Text Classification - Remaining Results}
\label{fig_text_rem}

Table~\ref{table:figurative_f1_remaining} shows the results for remaining models: Qwen3-8B, Gemma2-9B and GPT-OSS-20B. The same trend aligns: Idioms and metaphors have better classification score, while metonymy has a lower score. As mentioned in the main body, this is due to idioms and metaphors altering concreteness more than metonymy~\citep{lakoff1993contemporary, cacciari1993idioms, barcelona123}. The consistent results across different LLM families indicate that the concreteness axis can be universally used for figurative text classification.

\begin{table*}[t]
\centering
\resizebox{0.98\linewidth}{!}{
\begin{tabular}{llcccccc}
\toprule
 &  & \multicolumn{2}{c}{\textbf{Qwen}} & \multicolumn{2}{c}{\textbf{Gemma}} & \multicolumn{2}{c}{\textbf{GPT}} \\
\cmidrule(lr){3-4} \cmidrule(lr){5-6} \cmidrule(lr){7-8}
\textbf{Task} & \textbf{Dataset} & \textbf{Subspace} &  \textbf{Full Rep.} & \textbf{Subspace} &  \textbf{Full Rep.} & \textbf{Subspace} &  \textbf{Full Rep.} \\
\midrule
\multirow{2}{*}{Idioms} 
    & MAGPIE & 94.3 & 97.2 & 97.4 & 98.7 & 93.6 & 98.8  \\
    &  EPIE & 93.1 & 97.9 & 97.6 & 98.6 & 91.7 & 98.5 \\
\midrule
\multirow{2}{*}{Metaphor} 
    & VUA & 93.1 & 97.0 & 96.7 & 98.8 & 94.1 & 98.8 \\
    & MUNCH & 92.0 & 96.3  & 96.5 & 98.6 & 93.2 & 97.9 \\
\midrule
\multirow{2}{*}{Metonymy} 
    & ConMeC & 55.4  & 59.4 & 61.2 & 61.5 & 59.3 & 66.5 \\
    & MetFuse &  85.5  & 97.0  & 86.2 & 98.2 & 85.3 & 97.5 \\
\bottomrule
\end{tabular}
}
\caption{AUROC scores for figurative text classification using Qwen3-8B, Gemma2-9B and GPT-OSS-20B comparing a unidirectional concreteness subspace learned from Wikipedia and applied to downstream datasets against a full-representation classifier trained separately on each dataset.}
\label{table:figurative_f1_remaining}
\end{table*}

\section{Computational Resources}

Computational resources are also an important factor in this work. Our analysis requires extracting layer-wise embeddings from large LLMs and running plenty of probing operations, which is computationally demanding. We rely heavily on GPU acceleration (multiple NVIDIA A100 GPUs) as well as substantial CPU time for mathematical operations such as DiffMean analysis. Runtime is a practical constraint: for example, layer-wise MLP probing with 10-fold cross-validation on larger models can take several hours to complete.

\end{document}